\documentclass{article}

\usepackage[preprint]{nips_2018}

\usepackage[utf8]{inputenc} 
\usepackage[T1]{fontenc}    
\usepackage{hyperref}       
\usepackage{url}            
\usepackage{booktabs}       
\usepackage{amsfonts}       
\usepackage{nicefrac}       
\usepackage{microtype}      

\usepackage{graphicx}
\usepackage{caption}
\usepackage{subcaption}
\usepackage{tikz}
\usepackage{natbib}

\title{Super-Convergence: Very Fast Training of Neural Networks Using Large Learning Rates}

\author{Leslie N.~Smith \\
	U.S. Naval Research Laboratory, Code 5514\\
	4555 Overlook Ave., SW., Washington, D.C.  20375\\
	\texttt{ leslie.smith@nrl.navy.mil}
	\And
	Nicholay Topin \\
	University of Maryland, Baltimore County \\
	Baltimore, MD 21250 \\
	\texttt{ntopin1@umbc.edu} \\
}

\begin{document}

\maketitle

\begin{abstract}
In this paper, we describe a phenomenon, which we named ``super-convergence'', where neural networks can be trained  an order of magnitude faster than with standard training methods. 
The existence of super-convergence is relevant to understanding why deep networks generalize well.
One of the key elements of super-convergence is training with one learning rate cycle and a large maximum learning rate.
A primary insight that allows super-convergence training is that large learning rates regularize the training, hence requiring a reduction of all other forms of regularization in order to preserve an optimal regularization balance.
We also derive a simplification of the Hessian Free optimization method to compute an estimate of the optimal learning rate.
Experiments demonstrate super-convergence for Cifar-10/100, MNIST and Imagenet datasets, and resnet, wide-resnet, densenet, and inception architectures.
In addition, we show that super-convergence provides a  greater boost in performance relative to standard training when the amount of labeled training data is limited.
The architectures to replicate this work will be made available upon publication.
\end{abstract}

	\section{Introduction}

While deep neural networks have achieved amazing successes in a range of applications, understanding why  stochastic gradient descent (SGD) works so well remains an open and active area of research.
Specifically, we show that, for certain hyper-parameter values, using very large learning rates with the cyclical learning rate (CLR) method \citep{smith2015no, smith2017cyclical} can speed up training by as much as an order of magnitude. 
We named this phenomenon ``super-convergence.''
In addition to the practical value of super-convergence training, this paper provides empirical support and theoretical insights to the active discussions in the literature on stochastic gradient descent (SGD) and understanding generalization.

\begin{figure} [tbh]
	\begin{subfigure}[b]{0.47\textwidth}
		\includegraphics[width=\textwidth]{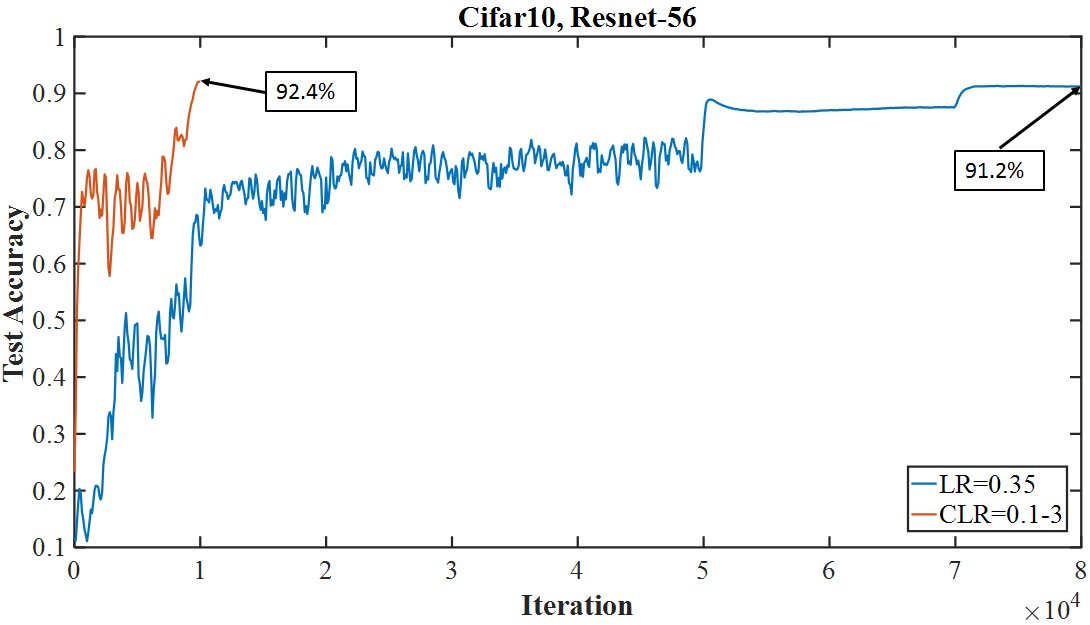}
		\caption{Comparison of test accuracies of super-convergence example to a typical (piecewise constant) training regime.}
		\label{fig:LRvsCLRResnet56}
	\end{subfigure}
	\quad
	\hfill
	~ 
	\centering
	\centering
	\begin{subfigure}[b]{0.47\textwidth}
		\includegraphics[width=\textwidth]{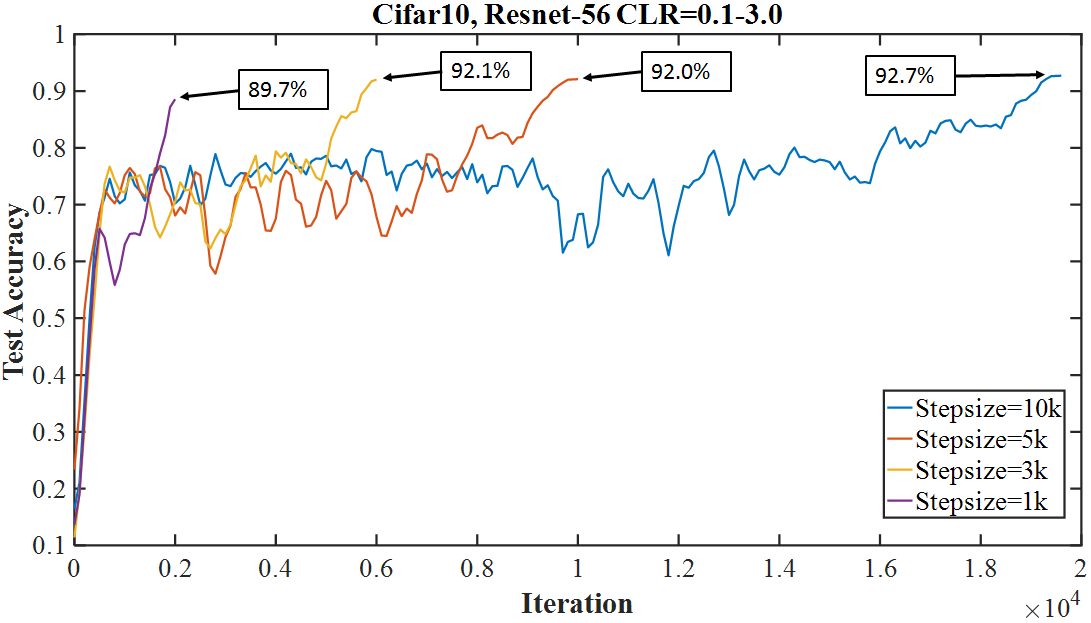}
		\caption{Comparison of test accuracies of super-convergence for a range of stepsizes.}
		\label{fig:clr3Resnet56}
	\end{subfigure}
	\caption{Examples of super-convergence with Resnet-56 on Cifar-10.}
	\label{fig:super}
	\vspace{-15pt}	
\end{figure}

Figure \ref{fig:LRvsCLRResnet56} provides a comparison of test accuracies from a super-convergence example and the result of a typical (piecewise constant) training regime for Cifar-10, both using a 56 layer residual network architecture.
Piecewise constant training reaches a peak accuracy of 91.2\% after approximately 80,000 iterations, while the super-convergence method reaches a higher accuracy (92.4\%) after only 10,000 iterations.
Note that the training accuracy curve for super-convergence is quite different than the characteristic accuracy curve (i.e., increasing, then plateau for each learning rate value). 
Figure \ref{fig:clr3Resnet56} shows the results for a range of CLR stepsize values, where training of one cycle reached a learning rate of 3.
This modified learning rate schedule achieves a higher final test accuracy (92.1\%) than typical training (91.2\%) after only 6,000 iterations.
In addition, as the total number of iterations increases from 2,000 to 20,000, the final accuracy improves from 89.7\% to 92.7\%.

The contributions of this paper include:
\begin{enumerate}
	\item Systematically investigates a new training methodology with improved speed and performance.
	\item Demonstrates that large learning rates regularize training and other forms of regularization must be reduced to maintain an optimal balance of regularization.
	\item Derives a simplification of the second order, Hessian-free optimization method to estimate optimal learning rates which demonstrates that large learning rates find wide, flat minima.
	\item Demonstrates that the effects of super-convergence are increasingly dramatic when less labeled training data is available.
\end{enumerate}

	\section{Background}
\label{sec:related}


In this paper, when we refer to a typical, standard, or a piecewise-constant training regime, it means the practice of using a global learning rate, (i.e., $\approx 0.1 $), for many epochs, until the test accuracy plateaus, and then continuing to train with a  learning rate decreased  by a factor of $ 0.1$.
This process of reducing the learning rate and continuing to train is often repeated two or three times.

There  exists extensive literature on stochastic gradient descent (SGD) (see \citet{goodfellow2016deep} and \citet{bottou2012stochastic}) which is relevant to this work.
Also, there exists a significant amount of literature on the loss function topology of deep networks (see \citet{chaudhari2016entropy}  for a review of the literature).
Our use of large learning rate values is in contrast to suggestions in the literature of a maximum learning rate value \cite{bottou2016optimization}.
Methods for adaptive learning rates have also been an active area of research.
This paper uses a simplification of the second order Hessian-Free optimization \citep{martens2010deep} to estimate optimal values for the learning rate.
In addition, we utilize some of the techniques described in \citet{schaul2013no} and \citet{gulcehre2017robust}.
Also, we show  that adaptive learning rate methods such as  Nesterov momentum \citep{sutskever2013importance,nesterov1983method}, AdaDelta \citep{duchi2011adaptive}, AdaGrad \citep{zeiler2012adadelta}, and Adam \citep{kingma2014adam} do not use sufficiently large learning rates when they are effective nor do they lead to super-convergence. 
A warmup learning rate strategy \citep{he2016deep,goyal2017accurate} could be considered a discretization of CLR, which was also recently suggested in \citep{jastrzkebski2017three}.
We  note that \citet{loshchilov2016sgdr} subsequently proposed a similar method to CLR, which they call SGDR.
The SGDR method uses a sawtooth pattern with a cosine followed by a jump back up to the original value.
Our experiments show that it is not possible to observe the super-convergence phenomenon when using their pattern.

Our work is intertwined with several active lines of research in the deep learning research community, including a lively discussion on stochastic gradient descent (SGD) and understanding why solutions generalize so well, research on SGD and the importance of noise for generalization, and the generalization gap between small and large mini-batches.
We defer our discussion of these lines of research to the Supplemental materials where we can compare to our empirical results and theoretical insights.


\begin{figure} [tbh]
	\begin{subfigure}[b]{0.4\textwidth}
		\includegraphics[width=\textwidth]{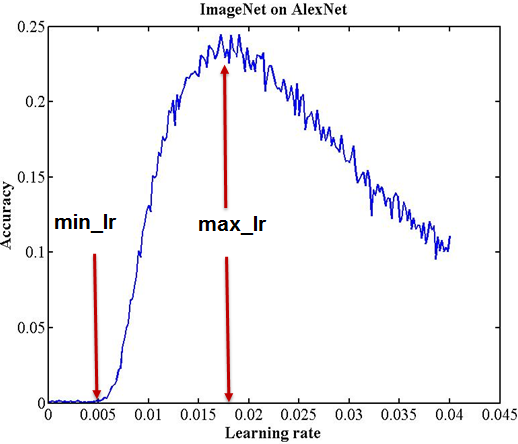}
		\caption{Typical learning rate range test result where there is a peak to indicate max\_lr.}
		\label{fig:normalRangeTest}
	\end{subfigure}
	\quad
	\hfill
	~ 
	\centering
	\centering
	\begin{subfigure}[b]{0.42\textwidth}
		\includegraphics[width=\textwidth]{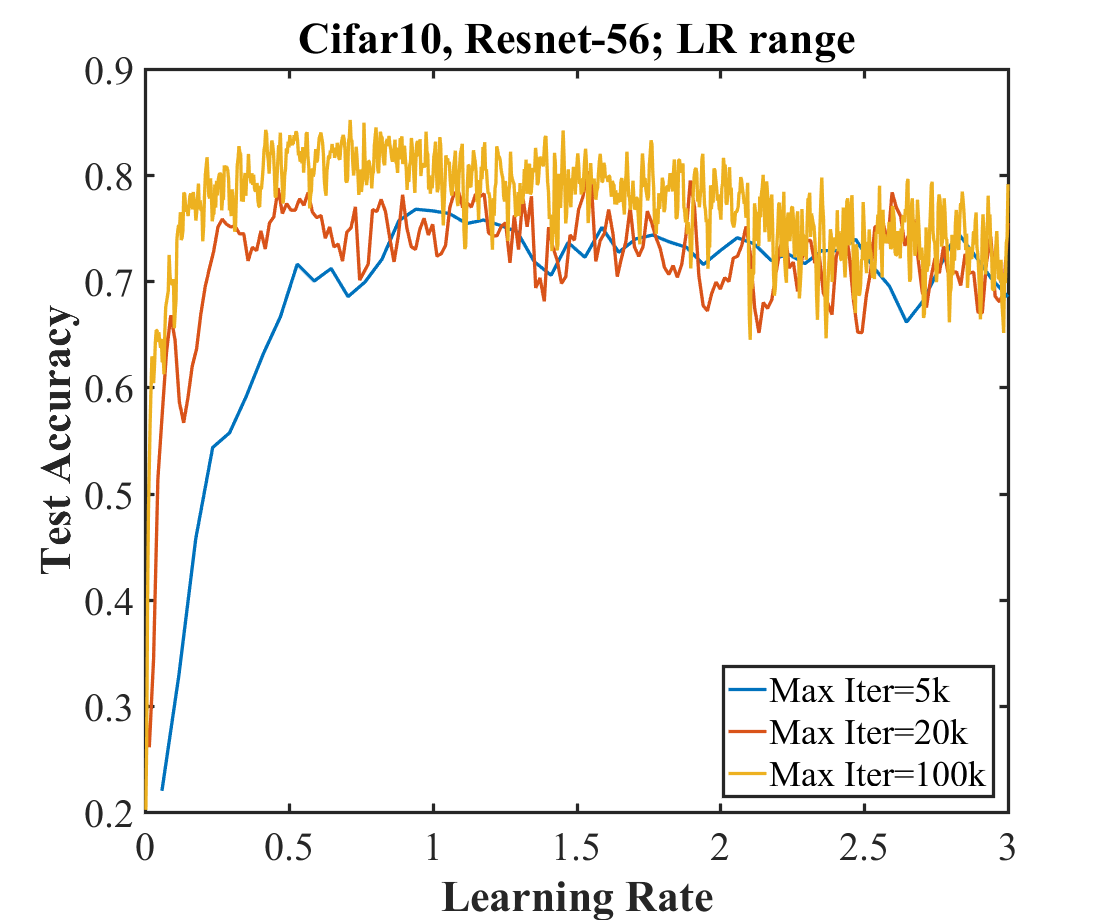}
		\caption{Learning rate range test result with the Resnet-56 architecture on Cifar-10. } 
		\label{fig:ResNetCifar10Range3}
	\end{subfigure}
	\caption{Comparison of learning rate range test results.}
	\label{fig:LRrange}
	\vspace{-10pt}
\end{figure}



\section{Super-convergence}
\label{sec:theory}

In this work, we use cyclical learning rates (CLR) and the learning rate range test (LR range test) which were first introduced by \citet{smith2015no} and later published in \citet{smith2017cyclical}.
To use CLR, one specifies minimum and maximum learning rate boundaries and a stepsize.
The stepsize is the number of iterations used for each step and a cycle consists of two such steps -- one in which the learning rate increases and the other in which it decreases.
\citet{smith2015no}  recommended the simplest method, which is letting the learning rate change linearly (on the other hand, \cite{jastrzkebski2017three} suggest discrete jumps).
Please note that the philosophy behind CLR is a \textbf{combination of curriculum learning  \citep{bengio2009curriculum} and simulated annealing \citep{aarts1988simulated}}, both of which have a long history of use in deep learning.

The LR range test  can be used to determine if super-convergence is possible for an architecture.
In the LR range test, training starts with a zero or very small learning rate which is slowly increased linearly throughout a  pre-training run.  
This provides information on how well the network can be trained over a range of learning rates.
Figure \ref{fig:normalRangeTest} shows a typical curve from a LR range test, where the test accuracy has a distinct peak.\footnote{Figure reproduced from \citet{smith2017cyclical} with permission.}
When starting with a small learning rate, the network begins to converge and, as the learning rate increases, it eventually becomes too large and causes the training/test accuracy to decrease.
The learning rate at this peak is the largest value to use as the maximum learning rate bound when using CLR.
The minimum learning rate can be  chosen by dividing the maximum by a factor of 3 or 4.
The optimal initial learning rate for a typical (piecewise constant) training regime usually falls between these minimum and maximum values. 

If one runs the LR range test for Cifar-10 on a 56 layer residual networks, one obtains the curves shown in Figure \ref{fig:ResNetCifar10Range3}.
Please note that learning rate values up to 3.0 were tested, which is an order of magnitude lager than typical values of the learning rate.
The test accuracy remains consistently high over this unusual range of large learning rates. 
This unusual behavior motivated our experimentation with much higher learning rates, and we believe that such behavior during a LR range test is indicative of potential for super-convergence.
The three curves in this figure are for runs with a maximum number of iterations of 5,000, 20,000, and 100,000, which shows independence between the number of iterations and the results. 

Here we suggest a slight modification of cyclical learning rate policy for super-convergence; always use one cycle that is smaller than the total number of iterations/epochs and allow the learning rate to decrease several orders of magnitude less than the initial learning rate for the remaining iterations.  We named this learning rate policy ``1cycle'' and in our experiments this policy allows an improvement in the accuracy.  

This paper shows that super-convergence traning can be applied universally and provides guidance on why, when and where this is possible. Specifically, there are many forms of regularization, such as large learning rates, small batch sizes, weight decay, and dropout \citep{srivastava2014dropout}.  Practitioners must balance the various forms of regularization for each dataset and architecture in order to obtain good performance.  
The general principle is: \emph{the amount of regularization must be balanced for each dataset and architecture.}  Recognition of this principle permits general use of super-convergence.  Reducing other forms of regularization and regularizing with very large learning rates makes training significantly more efficient.  


	\section{Estimating optimal learning rates}

Gradient or steepest descent is an optimization method that uses the slope as computed by the derivative to move in the direction of greatest negative gradient to iteratively update a variable.  
That is, given an initial point $x_0$, gradient descent proposes the next point to be:
\begin{equation}
x = x_0 - \epsilon \bigtriangledown_x f(x)
\label{eqn:LR}
\end{equation}
where $\epsilon$ is the step size or learning rate .  If we denote the parameters in a neural network (i.e., weights) as $ \theta \in R^N $ and $f(\theta)$ is the loss function, we can apply gradient descent to learn the weights of a network; i.e., with input $x$, a solution $y$, and non-linearity $\sigma$: 
\begin{equation}
y = f(\theta) = \sigma ( W_l \sigma  ( W_{l-1} \sigma ( W_{l-2} ... \sigma ( W_0 x + b_0) ... + b_l)
\end{equation}
where $W_l \in \theta$ are the weights for layer $l$ and $b_l \in \theta$ are biases for layer $l$.

The Hessian-free optimization method \citep{martens2010deep} suggests a second order solution that utilizes the slope information contained in the second derivative (i.e., the derivative of the gradient $\bigtriangledown_\theta f(\theta)$).
From \citet{martens2010deep},  the main idea of the second order Newton's method is that the loss function can be locally approximated by the quadratic as:
\begin{equation}
f(\theta) \approx f(\theta_0) + (\theta - \theta_0)^T \bigtriangledown_\theta f(\theta_0) + \frac{1}{2}  (\theta - \theta_0)^T H (\theta - \theta_0)
\label{eqn:2ndOrder}
\end{equation}
where $H$ is the Hessian, or the second derivative matrix of $f(\theta_0)$.
Writing Equation \ref{eqn:LR} to update the parameters at iteration $i$ as: 
\begin{equation}
\theta_{i+1} = \theta_{i} - \epsilon \bigtriangledown_\theta f(\theta_{i})
\label{eqn:theta}
\end{equation}
allows Equation \ref{eqn:2ndOrder} to be re-written as:
\begin{equation}
f(\theta_{i} - \epsilon \bigtriangledown_\theta f(\theta_{i})) \approx f(\theta_{i}) + (\theta_{i+1} - \theta_{i})^T \bigtriangledown_\theta f(\theta_{i}) + \frac{1}{2}  (\theta_{i+1} - \theta_{i})^T H (\theta_{i+1} - \theta_{i})
\end{equation}


In general it is not feasible to compute the Hessian matrix, which has $\Omega(N^2)$ elements, where $N$ is the number of parameters in the network, but it is unnecessary to compute the full Hessian.
The Hessian expresses the curvature in all directions in a high dimensional space, but the only relevant curvature direction is in the direction of steepest descent that SGD will traverse.
This concept is contained within Hessian-free optimization, as \citet{martens2010deep} suggests a finite difference approach for obtaining an estimate of the Hessian from two gradients:
\begin{equation}
H(\theta) = \lim_{\delta \to 0} \frac{\bigtriangledown f(\theta + \delta) - \bigtriangledown f(\theta)}{\delta}
\label{eqn:finite}
\end{equation}
where $\delta$ should be in the direction of the steepest descent.
The AdaSecant method \citep{gulcehre2014adasecant,gulcehre2017robust} builds an adaptive learning rate method based on this finite difference approximation as:
\begin{equation}
\epsilon^*  \approx \frac{\theta_{i+1} - \theta_{i} }{\bigtriangledown f(\theta_{i+1}) - \bigtriangledown f(\theta_{i})}
\label{eqn:LR2}
\end{equation}
where $ \epsilon^* $ represents the optimal learning rate for each of the neurons.
Utilizing Equation \ref{eqn:theta}, we rewrite Equation \ref{eqn:LR2} in terms of the differences between the weights from three sequential iterations as:
\begin{equation}
\epsilon^*  = \epsilon ~~ \frac{\theta_{i+1} - \theta_{i} }{2 \theta_{i+1}  - \theta_{i} - \theta_{i+2}}
\label{eqn:LR3}
\end{equation}
where $\epsilon$ on the right hand side is the learning rate value actually used in the  calculations to update the weights.
Equation \ref{eqn:LR3} is an expression for an adaptive learning rate for each weight update. 
We borrow the method in \citet{schaul2013no} to obtain an estimate of the global learning rate from the weight specific rates by summing over the numerator and denominator, with one minor difference.
In  \citet{schaul2013no} their expression is squared, leading to positive values -- therefore we sum the absolute values of each quantity to maintain positivity (using the square root of the sum of squares of the numerator and denominator of Equation \ref{eqn:LR3} leads to similar results).

\begin{figure} [tbh]
	\begin{subfigure}[b]{0.47\textwidth}
		\includegraphics[width=\textwidth]{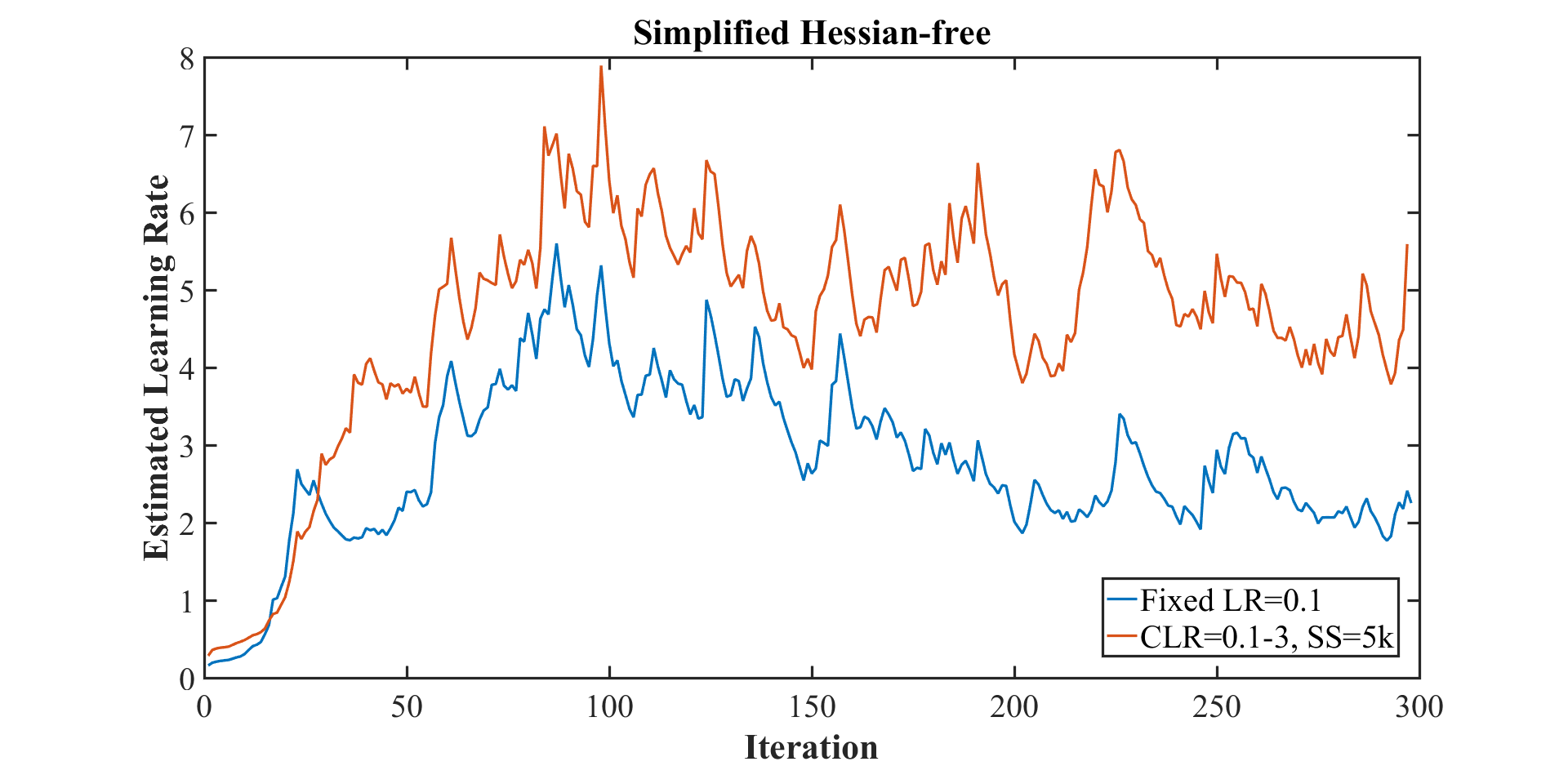}
		\caption{Estimated learning rates computed from the weights at every iteration for 300 iterations.}
		\label{fig:estLR300}
	\end{subfigure}
	\quad
	\hfill
	~ 
	\centering
	\begin{subfigure}[b]{0.47\textwidth}
		\includegraphics[width=\textwidth]{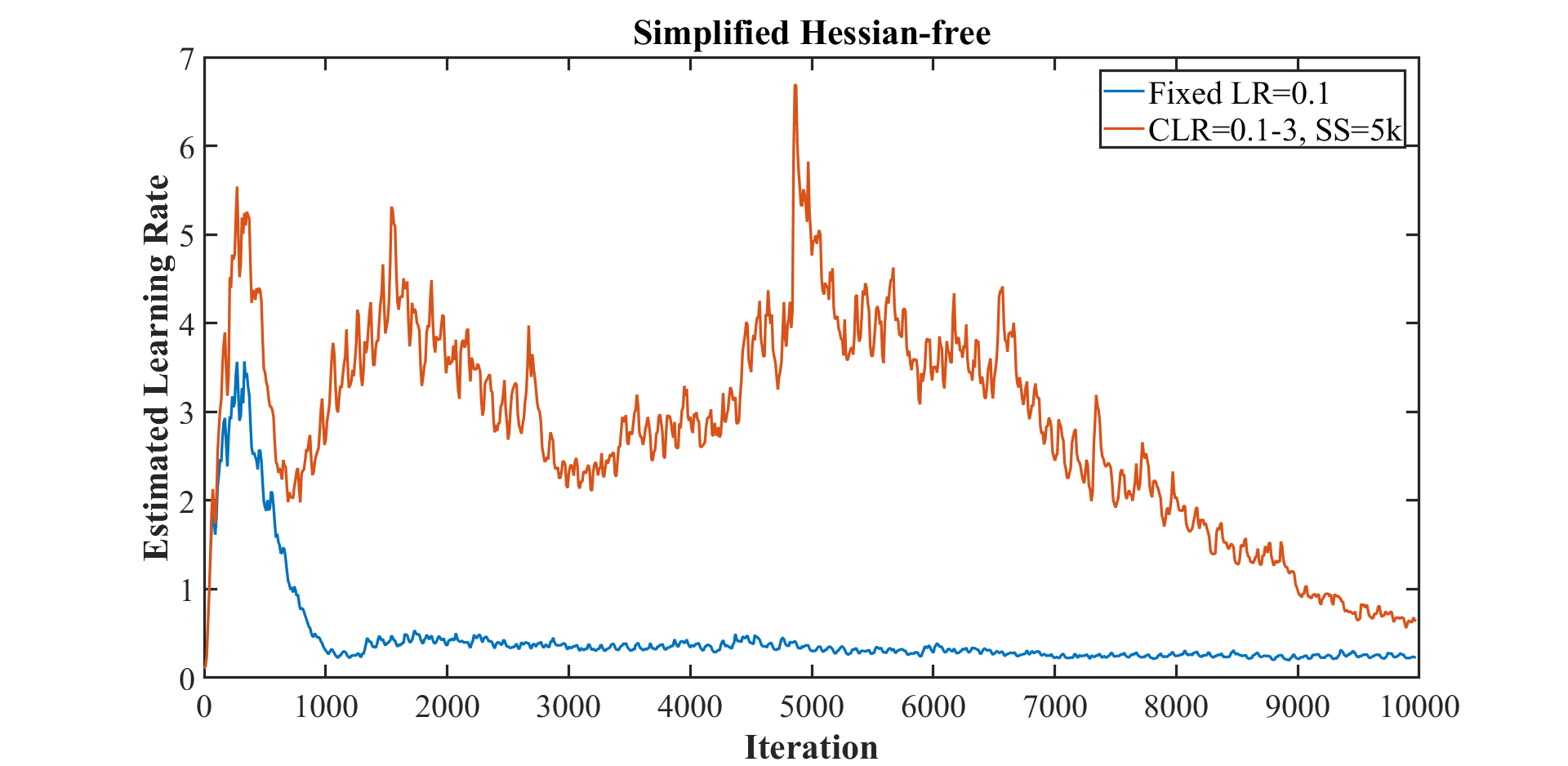}
		\caption{Estimated learning rates computed from the weights at every 10 iterations for 10,000 iterations.}
		\label{fig:estLR10k}
	\end{subfigure}
	\centering
	\caption{Estimated learning rate from the simplified Hessian-free optimization while training. The computed optimal learning rates are in the range from 2 to 6.}  
	\label{fig:estLR}
	\vspace{-10pt}
\end{figure}

For illustrative purposes, we computed the optimal learning rates from the weights of every iteration using Equation \ref{eqn:LR3} for two runs: first when the learning rate was a constant value of  $0.1$ and second with CLR in the range of $0.1 - 3$ with a stepsize of 5,000 iterations.
Since the computed learning rate exhibited rapid variations, we computed a moving average of the estimated learning rate as $ LR = \alpha \epsilon^* + (1 - \alpha ) LR $ with $\alpha = 0.1 $  and the results are shown in Figure \ref{fig:estLR300}  for the first 300 iterations.
This curve qualitatively shows that the optimal learning rates should be in the range of 2 to 6 for this architecture.
In Figure \ref{fig:estLR10k}, we used the weights as computed every 10 iterations and ran the learning rate estimation for 10,000 iterations.  
An interesting divergence happens here: when keeping the learning rate constant, the learning rate estimate initially spikes to a value of about 3 but then drops down near $0.2$.
On the other hand, the learning rate estimate using CLR remains high until the end where it settles down to a value of about $0.5$.
The large learning rates indicated by these Figures is caused by small values of our Hessian approximation and small values of the Hessian implies that SGD is finding flat and wide local minima.

In this paper we do not perform a full evaluation of the effectiveness of this technique as it is tangential to the theme of this work. 
We only use this method here to demonstrate that training with large learning rates are indicated by this approximation.
We leave a full assessment and tests of this method to estimate optimal adaptive learning rates as future work.




\begin{figure} [tbh]
	\begin{subfigure}[b]{0.47\textwidth}
		\includegraphics[width=\textwidth]{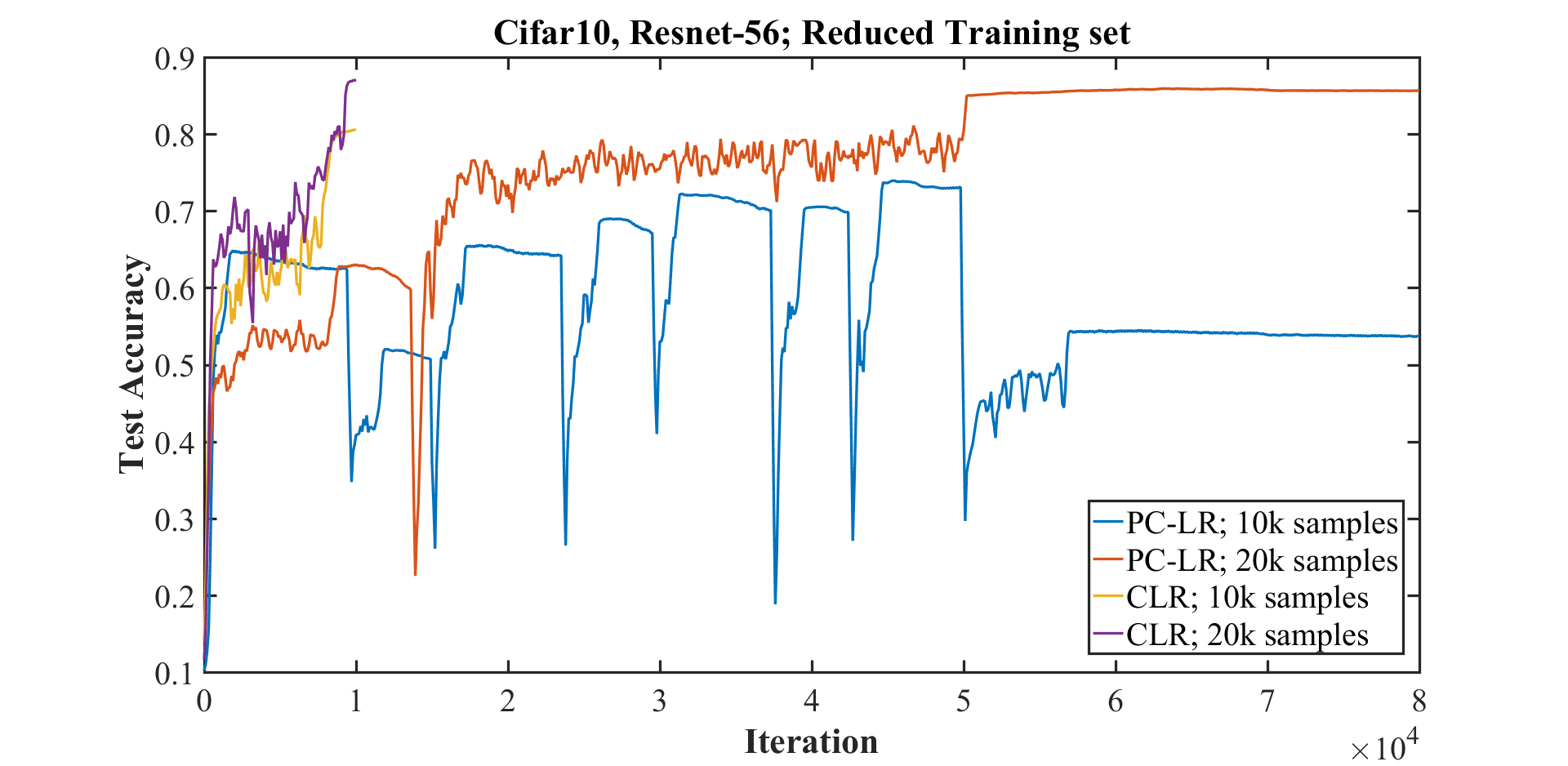}
		\caption{Comparison of test accuracies for Cifar-10/Resnet-56 with limited training samples.}
		\label{fig:CLRvsLRres56Cifar10-20k}
	\end{subfigure}
	\quad
	\hfill
	~ 
	\centering
	\centering
	\begin{subfigure}[b]{0.47\textwidth}
		\includegraphics[width=\textwidth]{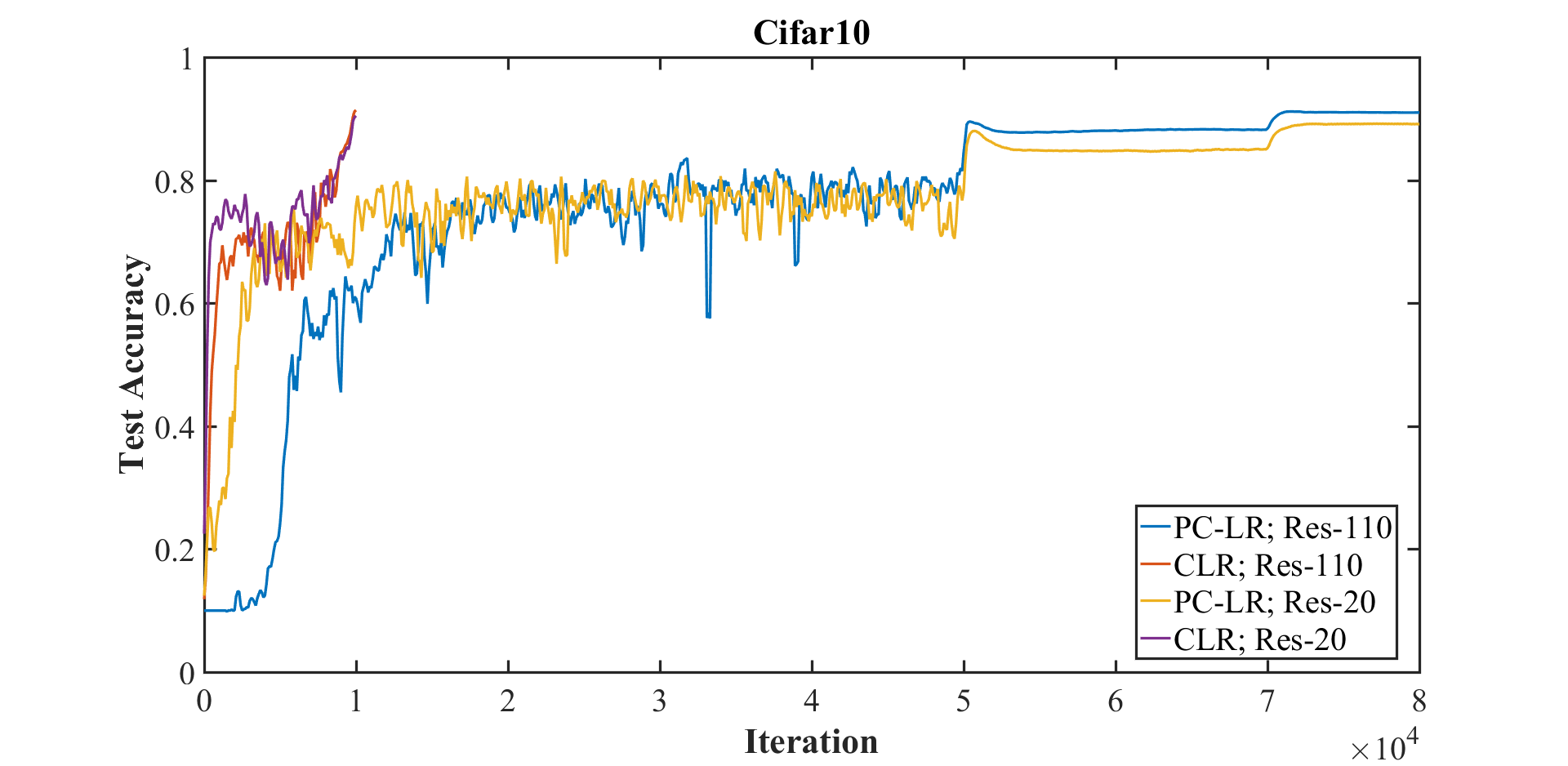}
		\caption{Comparison of test accuracies for Resnet-20 and Resnet-110.}
		\label{fig:LRvsCLRresnet20-110}
	\end{subfigure}
	\caption{Comparisons of super-convergence to typical training outcome with piecewise constant learning rate schedule. }
	\label{fig:SCWorks}
	\vspace{-10pt}
\end{figure}


\section{Experiments and analysis}
\label{sec:exp}

This section highlights a few of our more significant experiments. Additional experiments and details of our architecture are illustrated in the Supplemental Materials.  


\begin{table}[tb]
	\begin{center}
		\begin{tabular}{| c | c | c | c | c | }
			\hline
			\# training samples & Policy (Range)  & BN MAF &  Total Iterations &Accuracy (\%) \\ \hline
			40,000 & PC-LR=0.35  & 0.999 & 80,000 & 89.1 \\ \hline
			40,000 & CLR (0.1-3) & 0.95  & 10,000 & 91.1 \\ \hline
			30,000 & PC-LR=0.35  & 0.999 & 80,000 & 85.7 \\ \hline
			30,000 & CLR (0.1-3) & 0.95  & 10,000 & 89.6 \\ \hline
			20,000 & PC-LR=0.35  & 0.999 & 80,000 & 82.7 \\ \hline
			20,000 & CLR (0.1-3) & 0.95  & 10,000 & 87.9 \\ \hline
			10,000 & PC-LR=0.35  & 0.999 & 80,000 & 71.4 \\ \hline
			10,000 & CLR (0.1-3) & 0.95  & 10,000 & 80.6 \\ \hline
		\end{tabular}
		\vspace{10pt}
		\caption{Comparison of final accuracy results for various training regimes of Resnet-56 on Cifar-10. BN MAF is the value use for the $moving\_average\_fraction$ parameter with batch normalization. PC-LR is a standard piecewise constant learning rate policy described in Section \ref{sec:related} with an initial learning rate of 0.35.}
		\label{tab1:results}
	\end{center}
	\vspace{-25pt}
\end{table}



Figure \ref{fig:CLRvsLRres56Cifar10-20k} provides a  comparison of super-convergence with a reduced number of training samples.
When the amount of training data is limited, the gap in performance between the result of standard training and super-convergence increases.
Specifically, with a  piecewise constant learning rate schedule  the training encounters difficulties and diverges along the way.
On the other hand, a network trained with specific CLR parameters exhibits the super-convergence training curve and trains without difficulties.
The highest accuracies attained using standard learning rate schedules are listed  in Table \ref{tab1:results} and super-convergence test accuracy is 1.2\%, 5.2\%, and 9.2\% better for 50,000, 20,000, and 10,000 training cases, respectively.  
Hence, super-convergence becomes more beneficial when  training data is more limited.

We also ran experiments with Resnets with a number of layers in the range of 20 to 110 layers; that is, we ran experiments on residual networks with $l$ layers, where $ l = 20 + 9 n;$ for $ n = 0, 1, ... 10$.
Figure \ref{fig:LRvsCLRresnet20-110} illustrates the results for Resnet-20 and Resnet-110, for both a typical (piecewise constant) training regime with a standard initial learning rate of 0.35 and for CLR with a stepsize of 10,000 iterations.
The accuracy increase due to super-convergence is greater for the shallower architectures (Resnet-20: CLR 90.4\% versus piecewise constant LR schedule 88.6\%) than for the deeper architectures (Resnet-110: CLR 92.1\% versus piecewise constant LR schedule 91.0\%).

\begin{figure} [tbh]
	\begin{subfigure}[b]{0.47\textwidth}
		\includegraphics[width=\textwidth]{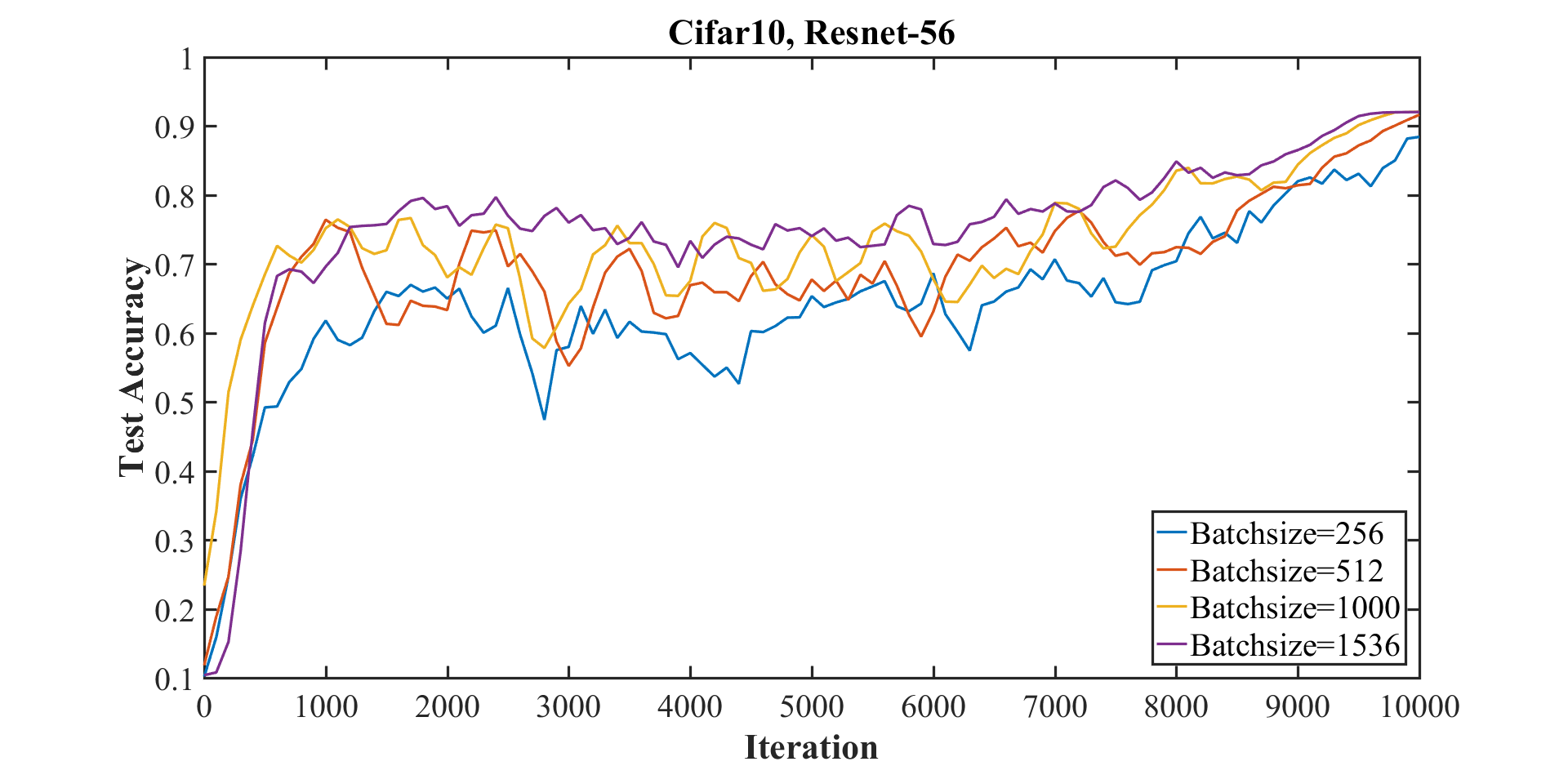}
		\caption{Comparison of test accuracies for Cifar-10, Resnet-56 for various total batch sizes.}
		\label{fig:clr3SS5kRes56BatchSizes7}
	\end{subfigure}
	\quad
	\hfill
	~ 
	\centering
	\begin{subfigure}[b]{0.47\textwidth}
		\includegraphics[width=\textwidth]{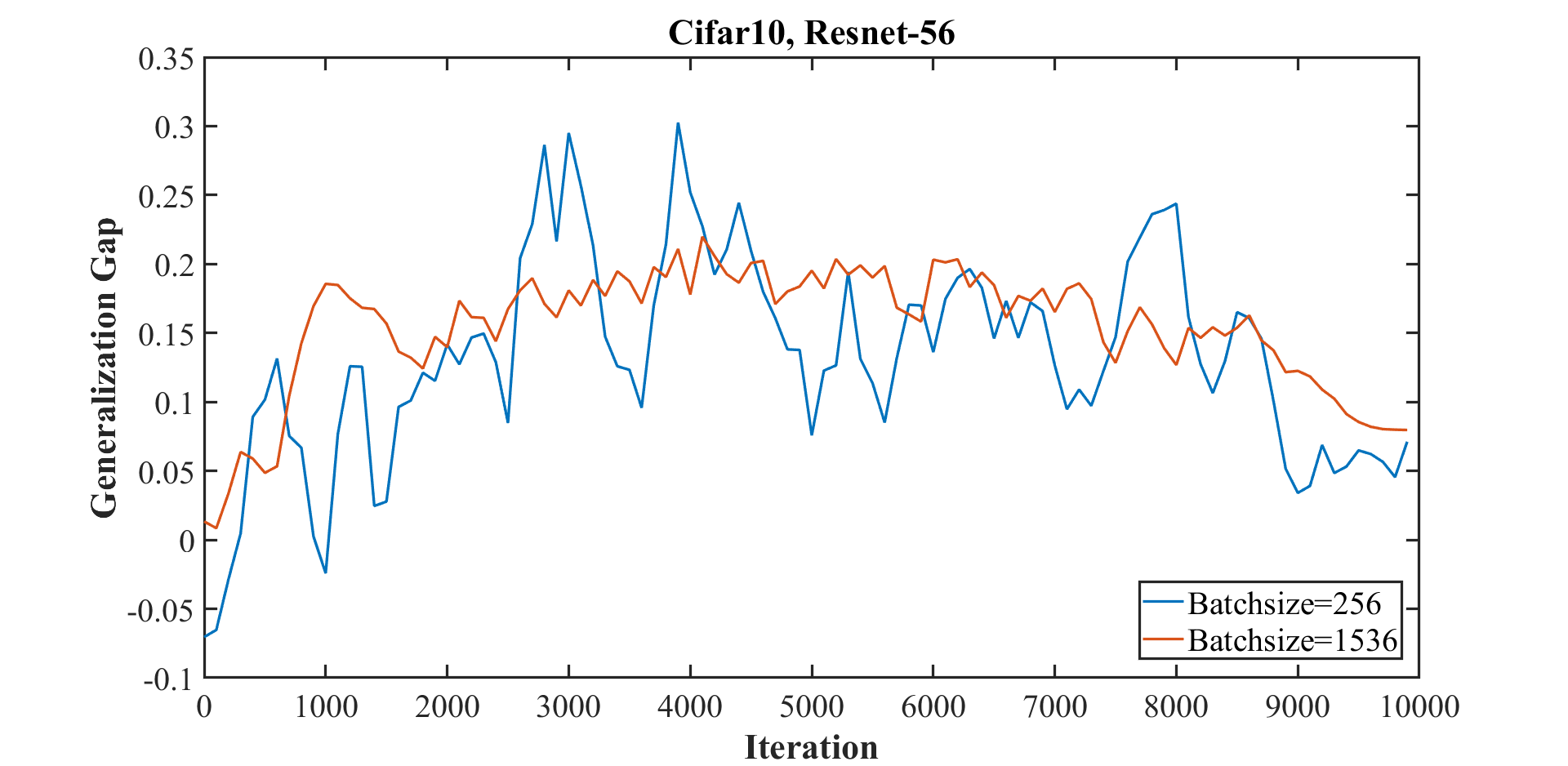}
		\caption{Comparison of the generalization gap (training - test accuracy) for a small and a large batch size.}
		\label{fig:clr3SS5kRes56BatchSizesGenGap}
	\end{subfigure}
	\centering
	\caption{Comparisons of super-convergence to over a range of batch sizes.  
		These results show that a large batch size is more effective than a small batch size for super-convergence training.}
	\label{fig:BatchSizes}
	\vspace{-5pt}	
\end{figure}

There are discussions in the deep learning literature on the effects of larger batch size and the generalization gap \citep{keskar2016large,jastrzkebski2017three,chaudhari2017stochastic,hoffer2017train}.
Hence, we investigated the effects total mini-batch size\footnote{Most of our reported results are with a total mini-batch size of 1,000 and we primarily used 8 GPUs and split the total mini-batch size 8 ways over the GPUs.} used in super-convergence training and found a small improvement in performance with larger batch sizes, as can be seen in Figure \ref{fig:clr3SS5kRes56BatchSizes7}.
In addition, Figure \ref{fig:clr3SS5kRes56BatchSizesGenGap} shows that the generalization gap (the difference between the training and test accuracies) are approximately equivalent for small and large mini-batch sizes.
This result differs than results reported elsewhere \citep{keskar2016large} and illustrates that a consequence of training with large batch sizes is the ability to use large learning rates.


\begin{table}[tb]
	\begin{center}
		\begin{tabular}{| c | c | c | c | c | c | c | }
			\hline
			Dataset & Architecture & CLR/SS/PL  & CM/SS & WD & Epochs & Accuracy (\%) \\ \hline
			Cifar-10 & wide resnet & 0.1/Step  & 0.9  & $10^{-4}$  &  100 & $ 86.7 \pm 0.6 $   \\ \hline
			Cifar-10 & wide resnet & 0.1/Step  & 0.9  & $10^{-4}$  &  200 & $ 88.7 \pm 0.6 $   \\ \hline
			Cifar-10 & wide resnet & 0.1/Step  & 0.9  & $10^{-4}$  &  400 & $ 89.8 \pm 0.4 $   \\ \hline
			Cifar-10 & wide resnet & 0.1/Step  & 0.9  & $10^{-4}$  &  800 & $ 90.3 \pm 1.0 $   \\ \hline
			Cifar-10 & wide resnet & 0.1-0.5/12  & 0.95-0.85/12 & $10^{-4}$ &  25 & $ 87.3 \pm 0.8 $   \\ \hline
			Cifar-10 & wide resnet & 0.1-1/23  & 0.95-0.85/23 & $10^{-4}$  &   50 & $ 91.3 \pm 0.1 $   \\ \hline
			Cifar-10 & wide resnet & 0.1-1/45  & 0.95-0.85/45 & $10^{-4}$  &  100 & $ 91.9 \pm 0.2 $   \\ \hline
			\hline   
			Cifar-10 & densenet & 0.1/Step  & 0.9 & $10^{-4}$  &  100 & $ 91.3 \pm 0.2 $   \\ \hline
			Cifar-10 & densenet & 0.1/Step  & 0.9 & $10^{-4}$  &  200 & $ 92.1 \pm 0.2 $   \\ \hline
			Cifar-10 & densenet & 0.1/Step  & 0.9 & $10^{-4}$  &  400 & $ 92.7 \pm 0.2 $   \\ \hline
			Cifar-10 & densenet & 0.1-4/22  & 0.9-0.85/22 & $10^{-6}$  &  50 & $ 91.7 \pm 0.3 $   \\ \hline
			Cifar-10 & densenet & 0.1-4/34  & 0.9-0.85/34 & $10^{-6}$  &  75 & $ 92.1 \pm 0.2 $   \\ \hline
			Cifar-10 & densenet & 0.1-4/45  & 0.9-0.85/45 & $10^{-6}$  &  100 & $ 92.2 \pm 0.2 $   \\ \hline
			Cifar-10 & densenet & 0.1-4/70  & 0.9-0.85/70 & $10^{-6}$  &  150 & $ 92.8 \pm 0.1 $   \\ \hline
			\hline   
			MNIST  & LeNet & 0.01/inv & 0.9  & $5 \times 10^{-4}$  &  85 & $ 99.03 \pm 0.04 $   \\ \hline
			MNIST  & LeNet & 0.01/step  & 0.9  & $5 \times 10^{-4}$  &  85 & $ 99.00 \pm 0.04 $   \\ \hline
			MNIST  & LeNet & 0.01-0.1/5  & 0.95-0.8/5 & $5 \times 10^{-4}$  &  12 & $ 99.25  \pm 0.03 $   \\ \hline
			MNIST  & LeNet & 0.01-0.1/12  & 0.95-0.8/12 & $5 \times 10^{-4}$  &  25 & $ 99.28  \pm 0.06 $   \\ \hline
			MNIST  & LeNet & 0.01-0.1/23  & 0.95-0.8/23 & $5 \times 10^{-4}$  &  50 & $ 99.27  \pm 0.07 $   \\ \hline
			MNIST  & LeNet & 0.02-0.2/40  & 0.95-0.8/40 & $5 \times 10^{-4}$  &  85 & $ 99.35  \pm 0.03 $   \\ \hline
			\hline   
			Cifar-100 & resnet-56 & 0.005/step & 0.9  & $10^{-4}$  &  100 & $ 60.8 \pm 0.4 $   \\ \hline
			Cifar-100 & resnet-56 & 0.005/step & 0.9  & $10^{-4}$  &  200 & $ 61.6 \pm 0.9 $   \\ \hline
			Cifar-100 & resnet-56 & 0.005/step & 0.9  & $10^{-4}$  &  400 & $ 61.0 \pm 0.2 $   \\ \hline
			Cifar-100 & resnet-56 & 0.1-0.5/12 & 0.95-0.85/12 & $10^{-4}$  &  25 & $ 65.4 \pm 0.2 $   \\ \hline
			Cifar-100 & resnet-56 & 0.1-0.5/23 & 0.95-0.85/23 & $10^{-4}$  &  50 & $ 66.4 \pm 0.6 $   \\ \hline
			Cifar-100 & resnet-56 & 0.09-0.9/45 & 0.95-0.85/45 & $10^{-4}$  &  100 & $ 69.0 \pm 0.4 $   \\ \hline
		\end{tabular}
		\vspace{10pt}
		\caption{Final accuracy and standard deviation for various datasets and architectures.  The total batch size (TBS) for all of the reported runs was 512.  PL = learning rate policy or SS = stepsize in epochs, where two steps are in a cycle, WD = weight decay, CM = cyclical momentum. Either SS or PL is provide in the Table and SS implies the cycle learning rate policy. }
		\label{tab:otherExamples}
	\end{center}
	\vspace{-20pt}
\end{table}

\subsection{Other datasets and architectures}
\label{other}

The wide resnet was created from a resnet with 32 layers by increasing the number of channels by a factor of 4 instead of the factor of 2 used by resnet.   Table \ref{tab:otherExamples} provides the final result of training using the 1cycle learning rate schedule with learning rate bounds from 0.1 to 1.0.  In 100 epochs, the wide32 network converges and provides a test accuracy of $91.9\% \pm 0.2$ while the standard training method achieves an accuracy of only $ 90.3 \pm 1.0 $ in 800 epochs.  This demonstrates super-convergence for wide resnets.

A 40 layer densenet architecture was create from the code at \url{https://github.com/liuzhuang13/DenseNetCaffe}. 
Table \ref{tab:otherExamples} provides the final accuracy results of training using the 1cycle learning rate schedule with learning rate bounds from 0.1 to 4.0 and cyclical momentum in the range of 0.9 to 0.85.  This Table also shows the effects of longer training lengths, where the final accuracy improves from 91.7\% for a quick 50 epoch (4,882 iterations) training to 92.8\% with a longer 150 epoch (14,648 iterations) training.  The step learning rate policy attains an equivalent accuracy of 92.7\% but requires 400 epochs to do so.

The MNIST database of handwritten digits from 0 to 9 (i.e., 10 classes)  has a training set of 60,000 examples, and a test set of 10,000 examples.  It is simpler than Cifar-10, so the shallow, 3-layer LeNet architecture was used for the tests.  Caffe \url{https://github.com/BVLC/caffe} provides the LeNet architecture and its associated hyper-parameters with the download of the Caffe framework. 
Table \ref{tab:otherExamples} lists the results of training the MNIST dataset with the LeNet architecture.  Using the hyper-parameters provided with the Caffe download, an accuracy of 99.03\% is obtained in 85 epochs.  Switching from an inv learning rate policy to a step policy produces equivalent results.  However, switching to the 1cycle policy produces an accuracy near 99.3\%, even in as few as 12 epochs.  

Cifar-100 is similar to CIFAR-10, except it has 100 classes and instead of 5,000 training images per class, there are only 500 training images and 100 testing images per class.
Table \ref{tab:otherExamples} compares the final accuracies of training with a step learning rate policy to training with a 1cycle learning rate policy.  The training results from the 1cycle learning rate policy are significantly higher than than the results from the step learning rate policy and the number of epochs required for training is greatly reduced (i.e., at only 25 epochs the accuracy is higher than any with a step learning rate policy).

\begin{figure} [tbh]
	\centering
	\begin{subfigure}[b]{0.47\textwidth}
		\includegraphics[width=\textwidth]{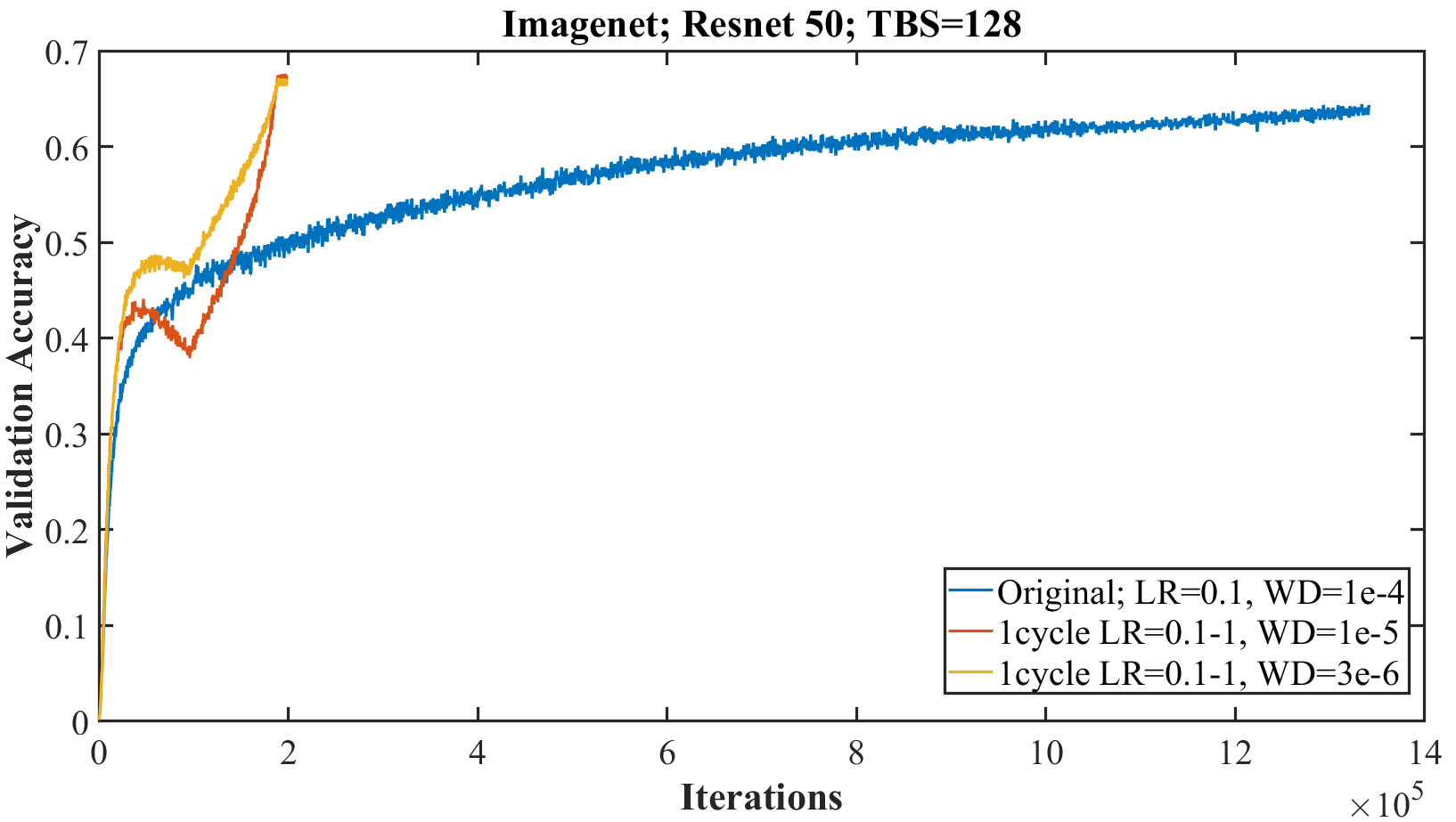}
		\caption{Resnet-50}
		\label{fig:imagenetResnetSC}       
	\end{subfigure}
	\quad
	\hfill
	~ 
	\centering
	\begin{subfigure}[b]{0.46\textwidth}
		\includegraphics[width=\textwidth]{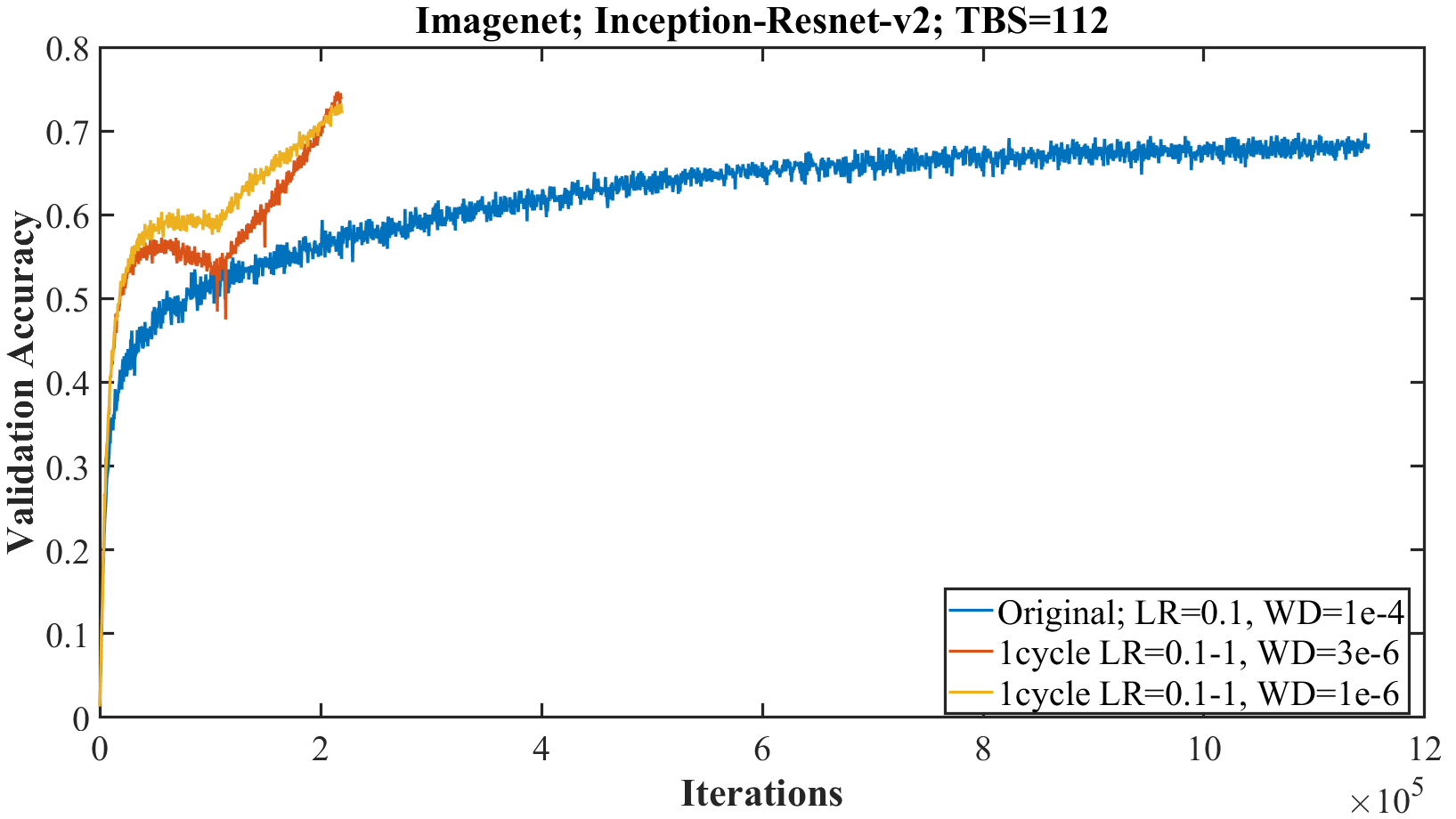}
		\caption{Inception-resnet-v2}
		\label{fig:imagenetInceptionSC}       
	\end{subfigure}
	\caption{Training resnet and inception architectures on the imagenet dataset with the standard learning rate policy (blue curve) versus a 1cycle policy that displays super-convergence.  Illustrates that deep neural networks can be trained much faster (20 versus 100 epochs) than by using the standard training methods.}
	\label{fig:imagenetResnet}
	\vspace{-5pt}	
\end{figure}

\subsection{Imagenet}
\label{sec:imagenet}

Our experiments show that reducing  regularization in the form of weight decay when training Imagenet allows the use of larger learning rates and produces much faster convergence and higher final accuracies.  

Figure \ref{fig:imagenetResnetSC} presents the comparison of training a resnet-50 architecture on Imagenet with the current standard training methodology versus the super-convergence method.  The hyper-parameters choices for the original training is set to the recommended values in \cite{szegedy2017inception} (i.e., momentum = 0.9, LR = 0.045 decaying every 2 epochs using an exponential rate of 0.94, WD = $10^{-4}$).  This produced the familiar training curve indicated by the blue lines in Figure \ref{fig:imagenetResnet}\footnote{Curves in the Figures for Imagenet are the average of only two runs.}.  The final top-1 test accuracy in Figure \ref{fig:imagenetResnetSC} is 63.7\% but the blue curve appears to be headed to an final accuracy near 65\%.

The red and yellow training curves in Figure \ref{fig:imagenetResnetSC} use the 1cycle learning rate schedule for 20 epochs, with the learning rate varying from 0.05 to 1.0, then down to 0.00005.  In order to use such large learning rates, it was necessary to reduce the value for weight decay.  Our tests found that weight decay values in the range from $3 \times 10^{-6}$ to $10^{-5}$ provided the best top-1 accuracy of $67.6\%$. Smaller values of weight decay showed signs of underfitting, which hurt performance  It is noteworthy that the two curves (especially for a weight decay value of $3 \times 10^{-6}$) display small amounts of overfitting.  Empirically this implies that small amounts of overfitting is a good indicator of the best value and helps the search for the optimal weight decay values early in the training.

Figure \ref{fig:imagenetInceptionSC} presents the training a inception-resnet-v2 architecture on Imagenet.  The blue curve is the current standard way while the red and yellow curves use the 1cycle learning rate policy.  The inception architecture tells a similar story as the resnet architecture.  Using the same step learning rate policy, momentum, and weight decay as in  \cite{szegedy2017inception} produces the familiar blue curve in Figure \ref{fig:imagenetInceptionSC}.  After 100 epochs the accuracy is 67.6\% and the curve can be extrapolated to an accuracy in the range of 69-70\%.

On the other hand, reducing the weight decay permitted use of the 1cycle learning rate policy with the learning rate varying from 0.05 to 1.0, then down to 0.00005 in 20 epochs.  The weight decay values in the range from $3 \times 10^{-6}$ to $10^{-6}$ work well and  using a weight decay of $3 \times 10^{-6}$ provides the best accuracy of $ 74.0\%$.  As with resnet-50, there appears to be a small amount of overfitting with this weight decay value. 

The lesson from these experiments is that deep neural networks can be trained much faster by super-convergence methodology than by the standard training methods.

\section{Conclusion}
\label{sec:conclusion}

We presented empirical evidence for a previously unknown phenomenon that we name super-convergence. In super-convergence, networks are trained with large learning rates in an order of magnitude fewer iterations and to a higher final test accuracy than when using a piecewise constant training regime.  Particularly noteworthy is the observation that the gains from super-convergence increase as the available labeled training data becomes more limited. Furthermore, this paper describes a simplification of the Hessian-free optimization method that we used for estimating learning rates. We demonstrate that super-convergence is possible with a variety of datasets and architectures, provided the regularization effects of large learning rates are balanced by reducing other forms of regularization.

We believe that a deeper study of super-convergence will lead to a better understanding of deep networks, their optimization, and their loss function landscape.

\bibliography{LargeLR.bib}
\bibliographystyle{plainnat}

	\appendix
	
	\section{Supplemental material}

	This Appendix contains information that does not fit within the main paper due to space restrictions.
	This includes an intuitive explanation for super-convergence, a discussion relating the empirical evidence from super-convergence to discussions in the literature, and the details of the experiments that were carried out in support of this research.

	\subsection{Intuitive explanation for super-convergence}
	\label{sec:intuitive}
	
\begin{figure} [tbh]
		\vspace{-5pt}
		\begin{subfigure}[b]{0.47\textwidth}
			\includegraphics[width=\textwidth]{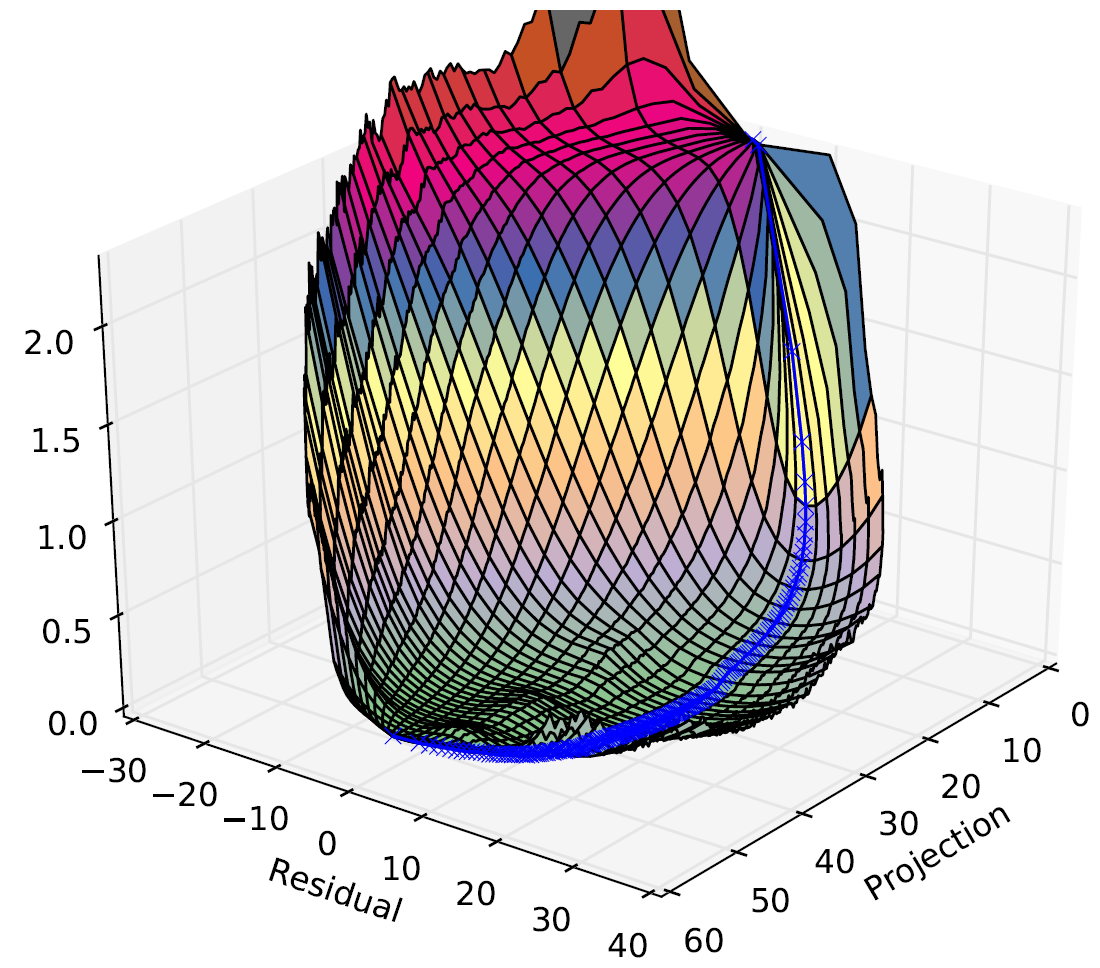}
			\caption{Visualization of how training transverses a loss function topology.}
			\label{fig:maxout_3d}
		\end{subfigure}
		\quad
		\hfill
		~ 
		\centering
		\centering
		\begin{subfigure}[b]{0.47\textwidth}
			\includegraphics[width=\textwidth]{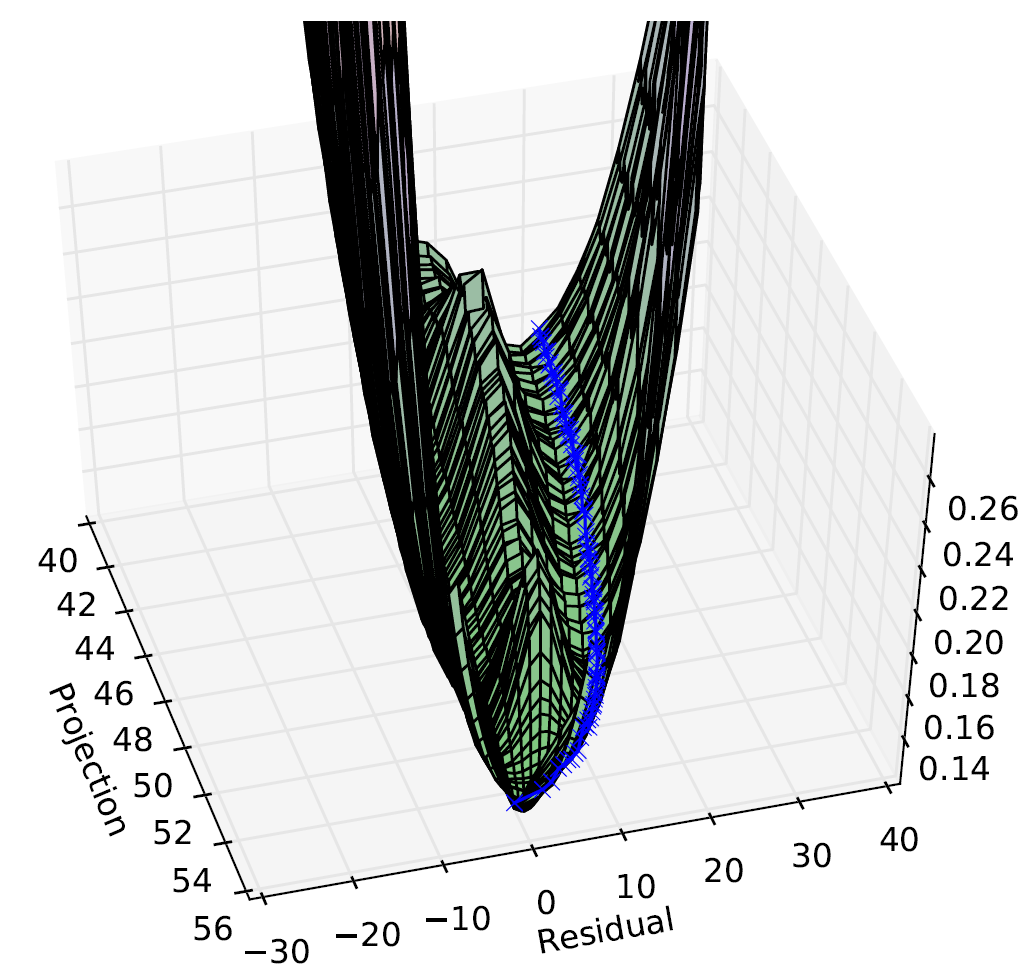}
			\caption{A close up of the end of the training for the example in Figure \ref{fig:maxout_3d}.} 
			\label{fig:maxout_3d_zoom}
		\end{subfigure}
		\caption{The 3-D visualizations from \cite{goodfellow2014qualitatively}.  The z axis represents the loss potential. }
		\label{fig:maxout}
		\vspace{-5pt}
	\end{figure}

	Figure \ref{fig:maxout_3d} provides an example of transversing the loss function topology.\footnote{Figure reproduced from \citet{goodfellow2014qualitatively} with permission.}
	This figure helps give an intuitive understanding of how super-convergence happens.
	The blue line in the Figure represents the trajectory of the training while converging and the x's indicate the location of the solution at each iteration and indicates the progress made during the training.
	In early iterations, the learning rate must be small in order for the training to make progress in an appropriate direction.
	The Figure also shows that significant progress is made in those early iterations.
	However, as the slope decreases so does the amount of progress per iteration and little improvement occurs over the bulk of the iterations.
	Figure \ref{fig:maxout_3d_zoom} shows a close up of the final parts of the training where the solution maneuvers through a valley to the local minimum within a trough.  
	
	Cyclical learning rates are well suited for training when the loss topology takes this form.  
	The learning rate initially starts small to allow convergence to begin.
	As the network traverses the flat valley, the learning rate is large, allowing for faster progress through the valley.
	In the final stages of the training, when the training needs to settle into the local minimum (as seen in Figure \ref{fig:maxout_3d_zoom}), the learning rate is once again reduced to a small value.

	
	\begin{figure} [tbh]
		\begin{subfigure}[b]{0.47\textwidth}
			\includegraphics[width=\textwidth]{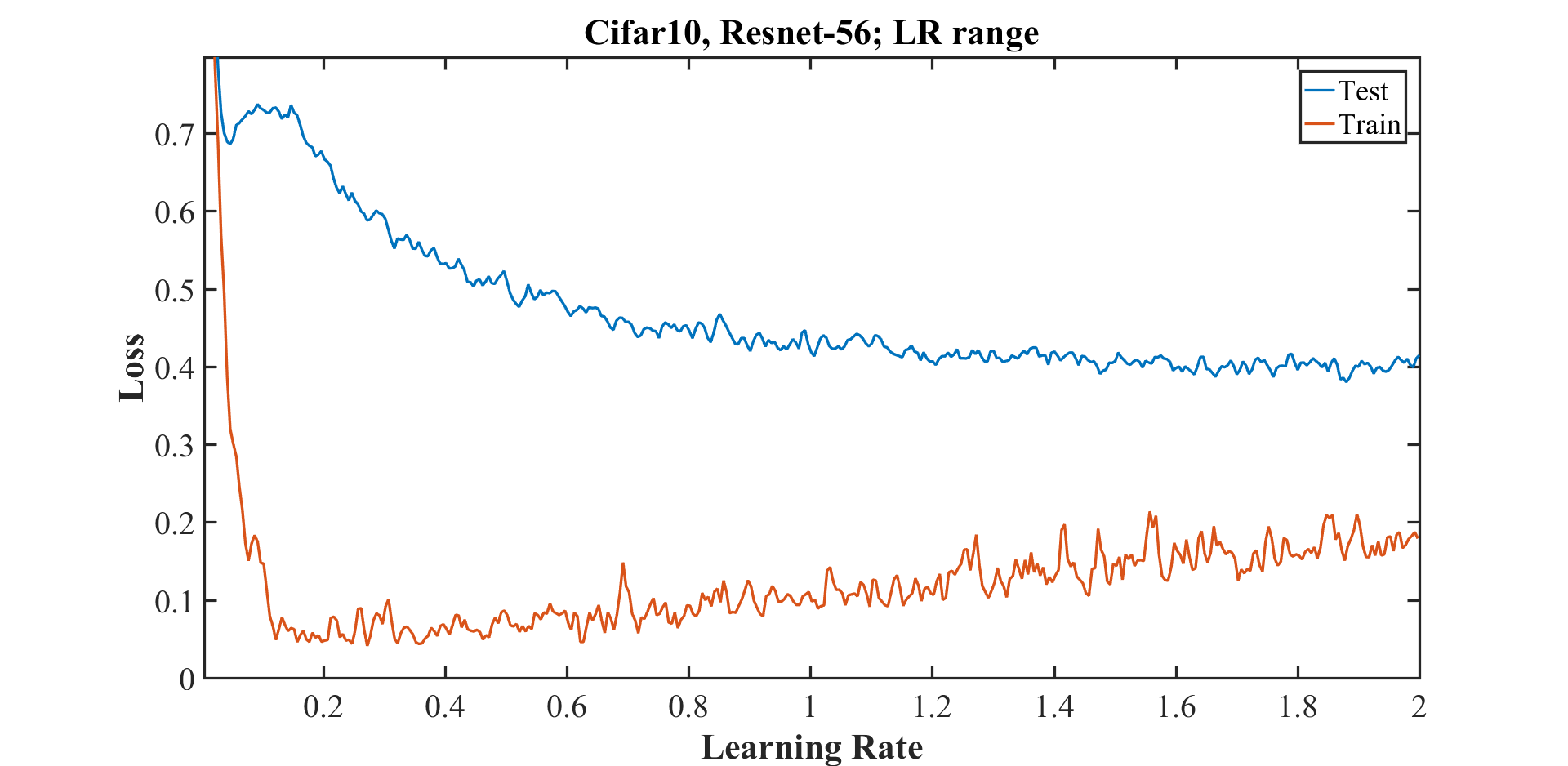}
			\caption{LR range test for Resnet-56.}
			\label{fig:regularizationRes56}
		\end{subfigure}
		\quad
		\centering
		\centering
		\begin{subfigure}[b]{0.47\textwidth}
			\includegraphics[width=\textwidth]{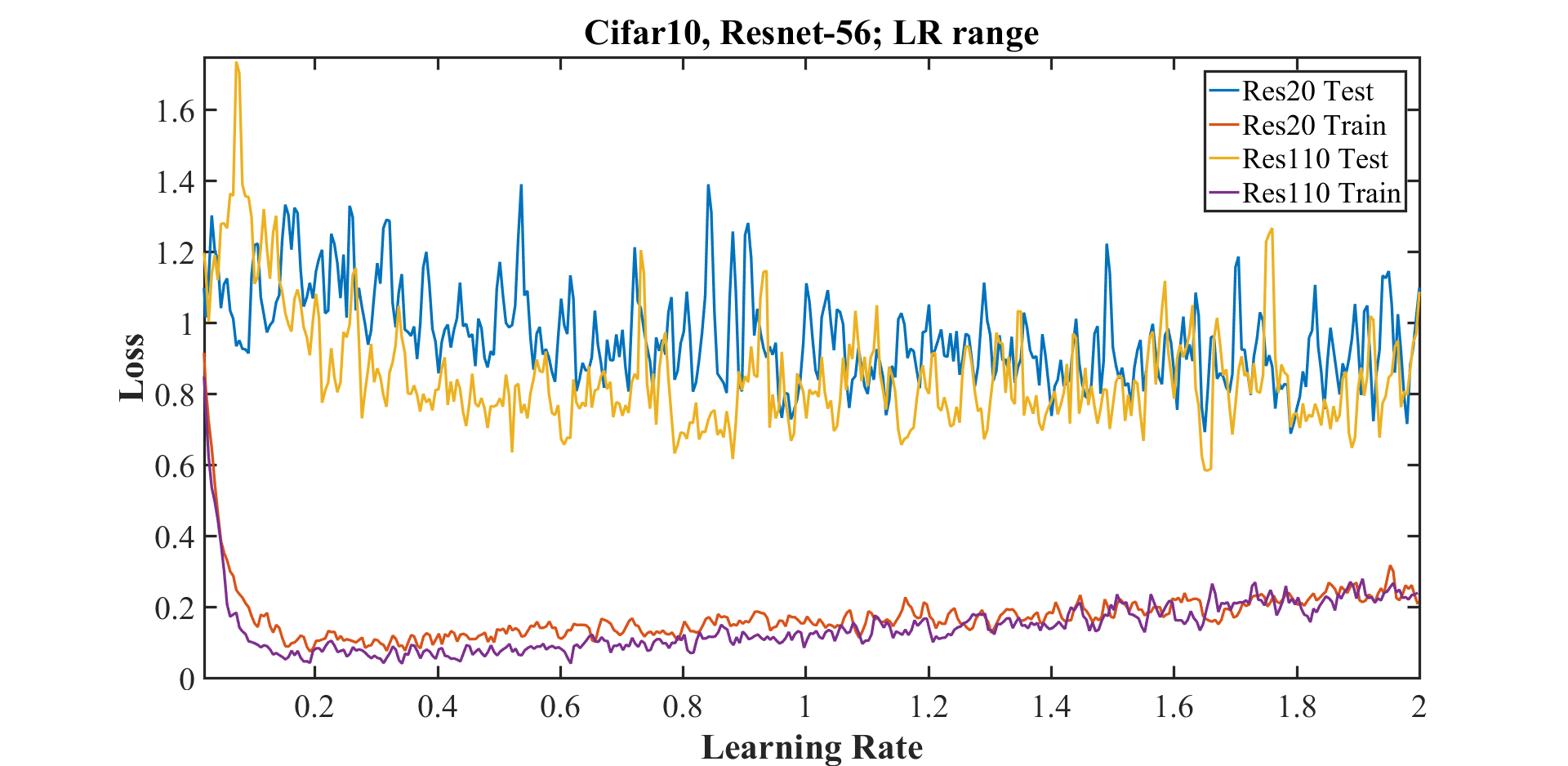}
			\caption{LR range tests for Resnet-20 and Resnet-110.}
			\label{fig:regularizationRes20-110}
		\end{subfigure}
		\caption{Evidence of regularization with large learning rates: decreasing generalization error as the learning rate increases from 0.3 to 1.5.}
		\label{fig:regCifar10ResNets}
		\vspace{-15pt}	
	\end{figure}
	

	\subsection{Large learning rate regularization}
	
	Super-convergence requires using a large maximum learning rate value.
	The LR range test reveals evidence of regularization through results shown in Figure \ref{fig:regularizationRes56}.
	Figure \ref{fig:regularizationRes56} shows  an increasing training loss and decreasing test loss while the learning rate increases from approximately 0.2 to 2.0 when training with the Cifar-10 dataset and a Resnet-56 architecture, which implies that regularization is occurring while training with these large learning rates.
	Similarly, Figure \ref{fig:regularizationRes20-110} presents the training and test loss curves for Resnet-20 and Resnet-110, where one can see the same decreasing generalization error.
	In addition, we ran the LR range test on residual networks with $l$ layers, where $ l = 20 + 9 n;$ for $ n = 0, 1, ... 10$ and obtained similar results.
	
	There is additional evidence that large learning rates are regularizing the training.
	As shown in the main paper, the final test accuracy results from a super-convergence training is demonstrably better than the accuracy results from a typical training method.
	Since a definition of regularization is ``any modification we make to a learning algorithm that is intended to reduce its generalization error'' \citep{goodfellow2016deep}, large learning rates should be considered as regularizing.
	Furthermore, others show that large learning rates leads to larger gradient noise, which leads to better generalization (i.e., \cite{jastrzkebski2017three,smith2017don}).
	
	
	\subsection{Relationship of super-convergence to SGD and generalization}
	\label{sec:relationship}

	There is substantial discussion in the literature on stochastic gradient descent (SGD) and understanding why solutions generalize so well (i.e, \cite{chaudhari2016entropy,chaudhari2017stochastic,im2016empirical,jastrzkebski2017three,smith2017understanding,kawaguchi2017generalization}).
	Super-convergence provides empirical evidence that supports some theories, contradicts some others, and points to the need for further theoretical understanding.
	We hope the response to super-convergence is similar to the reaction to the initial report of network memorization \citep{zhang2016understanding}, which sparked an active discussion within the deep learning research community on better understanding of the factors in SGD leading to solutions that generalize well (i.e., \citep{arpit2017closer}). 
	
	Our work impacts the line of research on SGD and the importance of noise for generalization.
	In this paper we focused on the use of CLR with very large learning rates, where large learning rates add noise/regularization in the middle of training.
	\cite{jastrzkebski2017three} stated that higher levels of noise lead SGD to solutions with better generalization.
	Specifically, they showed that the ratio of the learning rate to the batch size, along with the variance of the gradients, controlled the width of the local minima found by SGD.
	Independently, \cite{chaudhari2017stochastic} show that SGD performs regularization by causing SGD to be out of equilibrium, which is crucial to obtain good generalization performance, and derive that the ratio of the learning rate to batch size alone controls the entropic regularization term.
	They also state that data augmentation increases the diversity of SGD's gradients, leading to better generalization.
	The super-convergence phenomenon provides empirical support for these two  papers.
	
	In addition there are several other papers in the literature which state that wide, flat local minima produce solutions that generalize better than sharp minima \citep{hochreiter1997flat,keskar2016large,wu2017towards}.
	Our super-convergence results align with these results in the middle of training, yet a small learning rate is necessary at the end of training, implying that the minima of a local minimum is narrow. This needs to be reconciled.
	
	There are several papers on the generalization gap between small and large mini-batches and the relationship between gradient noise, learning rate, and batch size.
	Our results here supplements this other work by illustrating the possibility of time varying high noise levels during training.
	As mentioned above,  \cite{jastrzkebski2017three} showed that SGD noise is proportional to the learning rate, the variance of the loss gradients, divided by the batch size.
	Similarly Smith and Le \citep{smith2017understanding} derived the noise scale as $ g \approx \epsilon N / B (1 - m) $, where $g$ is the gradient noise, $\epsilon$ the learning rate, $N$ the number of training samples, and $m$ is the momentum coefficient.
	Furthermore, \cite{smith2017don} (also \cite{jastrzkebski2017three} independently) showed an equivalence of increasing batch sizes instead of a decreasing learning rate schedule.
	Importantly, these authors demonstrated that the noise scale $g$ is relevant to training and not the learning rate or batch size.
	Our paper proposes that the noise scale is a part of training regularization and that all forms of regularization must be balanced during training.
	
	\cite{keskar2016large} study the generalization gap between small and large mini-batches, stating that small mini-batch sizes lead to wide, flat minima and large batch sizes lead to sharp minima.
	They also suggest a batch size warm start for the first few epochs, then using a large batch size, which amounts to training with large gradient noise for a few epochs and then removing it.
	Our results contradicts this suggestion as we found it preferable to start training with little noise/regularization and let it increase (i.e., curriculum training), reach a noise peak, and reduce the noise level in the last part of training (i.e., simulated annealing). 
	
	\cite{goyal2017accurate} use a very large mini-batch size of up to 8,192 and adjust the learning rate linearly with the batch size.
	They also suggest a gradual warmup of the learning rate, which is a discretized version of CLR and matches our experience with an increasing learning rate.
	They make a point relevant to adjusting the batch size; if the network uses batch normalization, different mini-batch sizes leads to different statistics, which must be handled.
	\cite{hoffer2017train} made a similar point about batch norm and suggested using ghost statistics.
	Also, \cite{hoffer2017train}  show that longer training lengths is a form of regularization that improves generalization.
	On the other hand, we undertook a more comprehensive approach to all forms of regularization and our results demonstrate that regularization from training with very large learning rates permits much shorter training lengths.
	
	%
	
	Furthermore, this paper points the way towards new research directions, such as the following three:
	\begin{enumerate}
		\item \emph{Comprehensive investigations of regularization:} Characterizing ``good noise'' that improves the trained network's ability to generalize versus ``bad noise'' that interferes with finding a good solution (i.e., \cite{zhang2016understanding}).
		We find that there is a lack of a unified framework for treating SGD noise/diversity, such as architectural noise (e.g., dropout \citep{srivastava2014dropout}, dropconnect \citep{wan2013regularization}), noise from hyper-parameter settings (e.g., large learning rates and mini-batch sizes), adding gradient noise, adding noise to weights \citep{fortunato2017noisy} , and  input diversity (e.g., data augmentation, noise).
		\cite{smith2017understanding} took a step in this direction and \cite{smith2018disciplined} took another step  by studying the combined effects of learning rates, batch sizes, momentum, and weight decay.
		Gradient diversity has been shown to lead to flatter local minimum and better generalization.
		A unified framework should resolve conflicting claims in the literature on the value of each of these, such as for architectural noise (\cite{srivastava2014dropout} versus \cite{hoffer2017train}).
		Furthermore, many papers study each of these factors independently and by focusing on the trees, one might miss the forest.
		
		\item  \emph{Time dependent application of good noise during training:} As described above, combining curriculum learning with simulated annealing leads to cyclical application.
		To the best of our knowledge, this has only been applied sporadically in a few methods such as CLR \citep{,smith2017cyclical} or cyclical batch sizes \citep{jastrzkebski2017three} but these all fall under a single umbrella of time dependent gradient diversity (i.e., noise).
		Also, a method can be developed to learn an adaptive noise level while training the network.
		
		\item  \emph{Discovering new ways to stabilize optimization (i.e., SGD) with large noise levels:} Our evidence indicates that normalization (and batch normalization in particular) is a catalyst enabling super-convergence in the face of potentially  destabilizing noise from the large learning rates.  
		Normalization methods (batch norm, layer normalization \cite{ba2016layer}, streaming normalization \cite{liao2016streaming})  and new techniques (i.e., cyclical gradient clipping) to stabilize training need further investigation to discover better ways to keep SGD stable in the presence of enormous good noise.
	\end{enumerate}
	


	
	\begin{table}[tbh]
		\begin{center}
			\caption{A residual block which forms the basis of a Residual Network.}
			\begin{tabular}{| c | c | }
				\hline
				Layer & Parameters \\
				\hline \hline
				Conv Layer 1 & padding = 1 \\
				& kernel = 3x3 \\
				& stride = 1 \\
				& channels = $numChannels$ \\
				\hline
				Batch Norm 1 & moving\_average\_fraction=0.95 \\
				\hline
				ReLu 1 & --- \\
				\hline
				Conv Layer 2 & padding = 1 \\
				& kernel = 3x3 \\
				& stride = 1 \\
				& channels = $numChannels$ \\
				\hline
				Batch Norm 2 & moving\_average\_fraction=0.95 \\
				\hline
				Sum (BN2 output with original input) & --- \\
				\hline
				ReLu 2 & --- \\
				\hline
			\end{tabular}
			\label{table:resnet56_block}
		\end{center}
	\end{table}
	
	\begin{table}[tbh]
		\begin{center}
			\caption{A modified residual block which downsamples while doubling the number of channels.}
			\begin{tabular}{| c | c | }
				\hline
				Layer & Parameters \\
				\hline \hline
				Conv Layer 1 & padding = 1 \\
				& kernel = 3x3 \\
				& stride = 2 \\
				& channels = $numChannels$ \\
				\hline
				Batch Norm 1 & moving\_average\_fraction=0.95 \\
				\hline
				ReLu 1 & --- \\
				\hline
				Conv Layer 2 & padding = 1 \\
				& kernel = 3x3 \\
				& stride = 1 \\
				& channels = $numChannels$ \\
				\hline
				Batch Norm 2 & moving\_average\_fraction=0.95 \\
				\hline
				Average Pooling (of original input) & padding = 0 \\
				& kernel = 3x3 \\
				& stride = 2 \\
				\hline
				Sum (BN2 output with AvgPool output) & --- \\
				\hline
				ReLu 2 & --- \\
				\hline
				Concatenate (with zeroes) & channels = 2*$numChannels$ \\
				\hline
			\end{tabular}
			\label{table:resnet56_transition}
		\end{center}
	\end{table}
	
	\begin{table}[tbh]
		\begin{center}
			\caption{Overall architecture for ResNet-56. }
			\begin{tabular}{| c | c | c | }
				\hline
				Layer/Block Type & Parameters \\
				\hline \hline
				Conv Layer & padding = 1 \\
				& kernel = 3x3 \\
				& stride = 2 \\
				& channels = 16 \\
				\hline
				Batch Norm & moving\_average\_fraction=0.95 \\
				\hline
				ReLU & --- \\
				\hline
				ResNet Standard Block x9 & $numChannels$ = 16 \\
				\hline
				ResNet Downsample Block & $numChannels$ = 16 \\
				\hline
				ResNet Standard Block x8 & $numChannels$ = 32 \\
				\hline
				ResNet Downsample Block & $numChannels$ = 32 \\
				\hline
				ResNet Standard Block x8 & $numChannels$ = 64 \\
				\hline
				Average Pooling & padding = 0 \\
				& kernel = 8x8 \\
				& stride = 1 \\
				\hline
				Fully Connected Layer & --- \\
				\hline
				
			\end{tabular}
			\label{table:resnet56_arch}
		\end{center}
	\end{table}

	
	\subsection{Hardware, software, and architectures}
	
	All of the experiments were run with Caffe (downloaded October 16, 2016) using CUDA 7.0 and Nvidia's CuDNN.  
	Most of these experiments were run on a 64 node cluster with 8 Nvidia Titan Black GPUs, 128 GB memory, and dual Intel Xeon E5-2620 v2 CPUs per node and we utilized the multi-gpu implementation of Caffe.  The other experiments were run on an IBM Power8, 32 compute nodes, 20 cores/node, 4 Tesla P100 GPUs/node with 255 GB available memory/node.
	
	
	
	The Resnet-56 architecture consists of three stages. Within each stage, the same residual block structure is sequentially repeated. This structure is given in Table \ref{table:resnet56_block}. Between stages, a different residual block structure is used to reduce the spatial dimension of the channels. Table \ref{table:resnet56_transition} shows this structure. The overall architecture is described in Table \ref{table:resnet56_arch}. Following the Caffe convention, each Batch Norm layer was followed by a scaling layer to achieve true batch normalization behavior. 
	This and the other architectures necessary to replicate this work will be made available upon publication.
	
	Cifar-10 and Cifar-100 were downloaded from \url{https://www.cs.toronto.edu/~kriz/cifar.html}.  
	The MNIST database of handwritten digits  was obtained from \url{http://yann.lecun.com/exdb/mnist/}.
	The resnet architectures are available at \url{https://github.com/yihui-he/resnet-cifar10-caffe}.
	The wide resnet was created from a resnet with 32 layers by increasing the number of channels by a factor of 4 instead of the factor of 2 used by resnet.
	A 40 layer densenet architecture was create from the code at \url{https://github.com/liuzhuang13/DenseNetCaffe}.
	
	Imagenet, which is a large image database with 1.24 million training images for 1,000 classes, was downloaded from \url{http://image-net.org/download-images}.  The resnet-50 and inception-resnet-v2 architectures used for these experiments were obtained from \url{https://github.com/soeaver/caffe-model}.  
	
	\begin{figure} [tbh]
		\begin{subfigure}[b]{0.47\textwidth}
			\includegraphics[width=\textwidth]{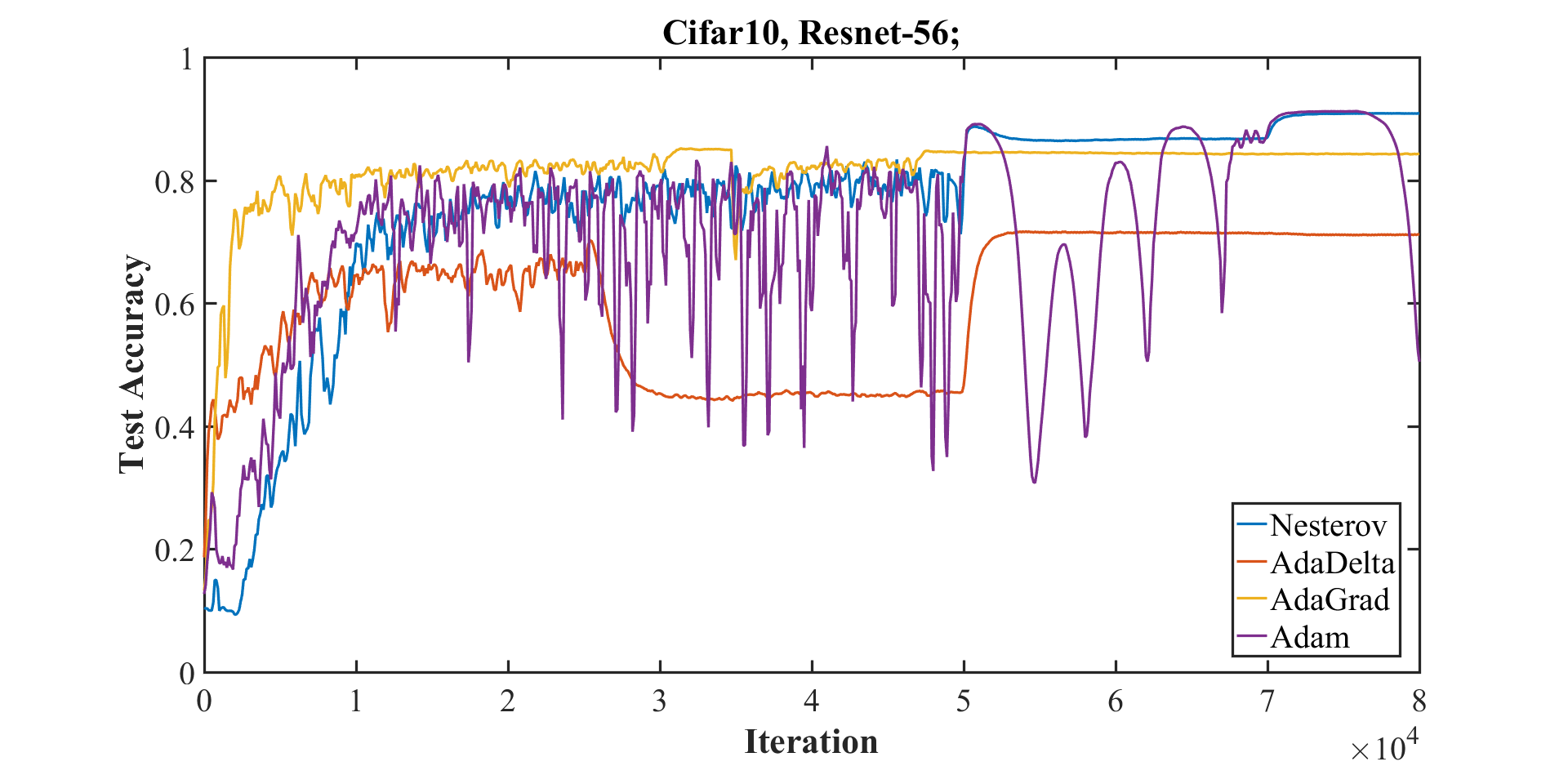}
			\caption{Comparison of test accuracies with various adaptive learning rate methods.}
			\label{fig:lr35Res56AdaptiveLR}
		\end{subfigure}
		\quad
		\hfill
		~ 
		\centering
		\centering
		\begin{subfigure}[b]{0.47\textwidth}
			\includegraphics[width=\textwidth]{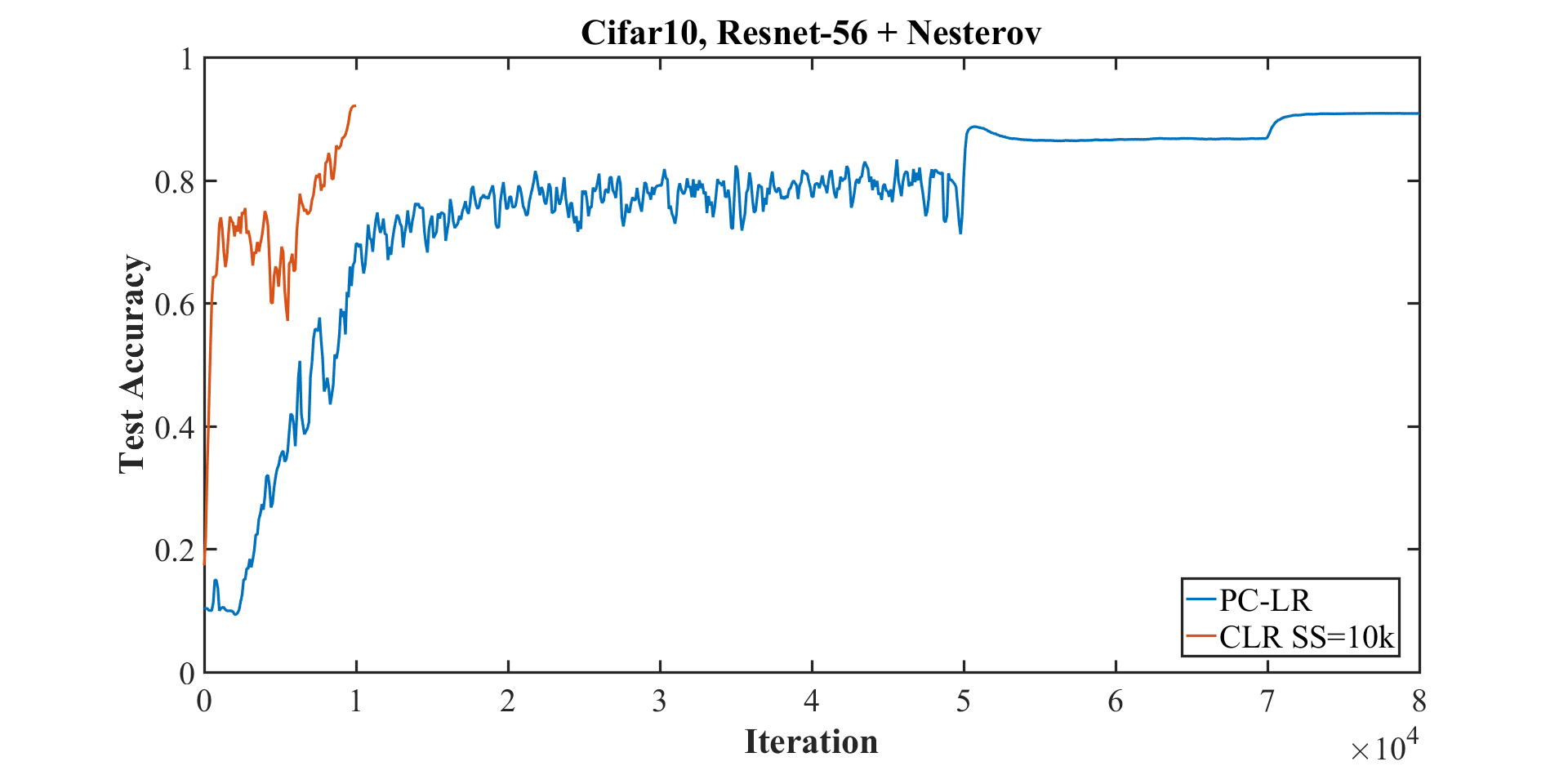}
			\caption{Comparison of test accuracies With Nesterov method.}
			\label{fig:clr3Res56Nesterov}
		\end{subfigure}
		\caption{Comparisons for Cifar-10, Resnet-56  of super-convergence to piecewise constant training regime. }
		\label{fig:adaptive}
	\end{figure}
	
	
	\subsection{Hyper-parameters and adaptive learning rate methods}
	\label{sec:HP}
	
	We tested the effect of batch normalization on super-convergence.
	Initially, we found that having $ use\_global\_stats: true $ in the test phase prevents super-convergence.
	However, we realized this was due to using the default value of $ moving\_average\_fraction = 0.999 $ that is only appropriate for the typical, long training times. 
	However, when a network is trained very quickly, as with super-convergence,  a value of $ 0.999 $ does not update the accumulated global statistics quickly enough.
	Hence, we found that the smaller values listed in Table \ref{tab1:results} (column BN MAF) were more appropriate.
	We found that these smaller values of $ moving\_average\_fraction$ worked well and we didn't need to resort to ghost statistics \citep{hoffer2017train}.
	
	
	We ran a series of experiments with a variety of ranges for CLR.
	Table \ref{tab1:results} shows the results for maximum learning rate bounds from 1.0 to 3.5.
	These experiments show that a maximum learning rate of approximately 3 performed well.
	
	We also investigated whether adaptive learning rate methods in a piecewise constant training regime would learn to adaptively use large learning rates to improve training speed and performance.
	We tested  Nesterov momentum \citep{sutskever2013importance,nesterov1983method}, AdaDelta \citep{duchi2011adaptive}, AdaGrad \citep{zeiler2012adadelta}, and Adam \citep{kingma2014adam} on Cifar-10 with the Resnet-56 architecture but none of these methods speed up the training process in a similar fashion to super-convergence.
	Specifically, Figure \ref{fig:lr35Res56AdaptiveLR} shows the results of this training for Nesterov momentum \citep{sutskever2013importance,nesterov1983method}, AdaDelta \citep{duchi2011adaptive}, AdaGrad \citep{zeiler2012adadelta}, and Adam \citep{kingma2014adam}.
	We found no sign that any of these methods discovered the utility of large learning rates nor any indication of super-convergence-like behavior.
	This is important because this lack is indicative of the failing in the theory behind these approaches and imply that there is a need for further work on adaptive methods.
	
	We also ran these adaptive learning methods with the 1cycle learning rate policy and found that  Nesterov, AdaDelta, and AdaGrad allowed super-convergence to occur, but we were unable to create this phenomenon with Adam.
	For example, Figure \ref{fig:clr3Res56Nesterov} shows a comparison of super-convergence to a piecewise constant training regime with the Nesterov momentum method.
	Here super-convergence yields a final test accuracy after 10,000 iterations of 92.1\% while the piecewise constant training regime at iteration 80,000 has an accuracy of 90.9\%.

	
	\begin{table}[tb]
		\begin{center}
			\begin{tabular}{| c | c | c | c | c | }
				\hline
				\# training samples & Policy (Range)  & BN MAF &  Total Iterations &Accuracy (\%) \\ \hline
				50,000 & CLR (0.1-3.5) & 0.95 & 10,000 & 92.1 \\ \hline
				50,000 & CLR (0.1-3)   & 0.95 & 10,000 & 92.4 \\ \hline
				50,000 & CLR (0.1-2.5) & 0.95 & 10,000 & 92.3 \\ \hline  
				50,000 & CLR (0.1-2)   & 0.95 & 10,000 & 91.7 \\ \hline
				50,000 & CLR (0.1-1.5) & 0.95 & 10,000 & 90.9 \\ \hline
				50,000 & CLR (0.1-1)   & 0.95 & 10,000 & 91.3 \\ \hline
				\hline
				50,000 & CLR (0.1-3)   & 0.97 & 20,000 & 92.7 \\ \hline
				50,000 & CLR (0.1-3)   & 0.95 & 10,000 & 92.4 \\ \hline
				50,000 & CLR (0.1-3)   & 0.93 & 8,000 & 91.7 \\ \hline
				50,000 & CLR (0.1-3)   & 0.90 & 6,000 & 92.1 \\ \hline
				50,000 & CLR (0.1-3)   & 0.85 & 4,000 & 91.1 \\ \hline
				50,000 & CLR (0.1-3)   & 0.80 & 2,000 & 89.7 \\ \hline
			\end{tabular}
			\vspace{10pt}
			\caption{Comparison of final accuracy results for various training regimes of Resnet-56 on Cifar-10. BN MAF is the value use for the $moving\_average\_fraction$ parameter with batch normalization. PC-LR is a standard piecewise constant learning rate policy described in Section \ref{sec:related} with an initial learning rate of 0.35.}
			\label{tab1:results}
		\end{center}
		\vspace{-25pt}
	\end{table}
	

	
	\begin{figure} [tbh]
		\begin{subfigure}[b]{0.47\textwidth}
			\includegraphics[width=\textwidth]{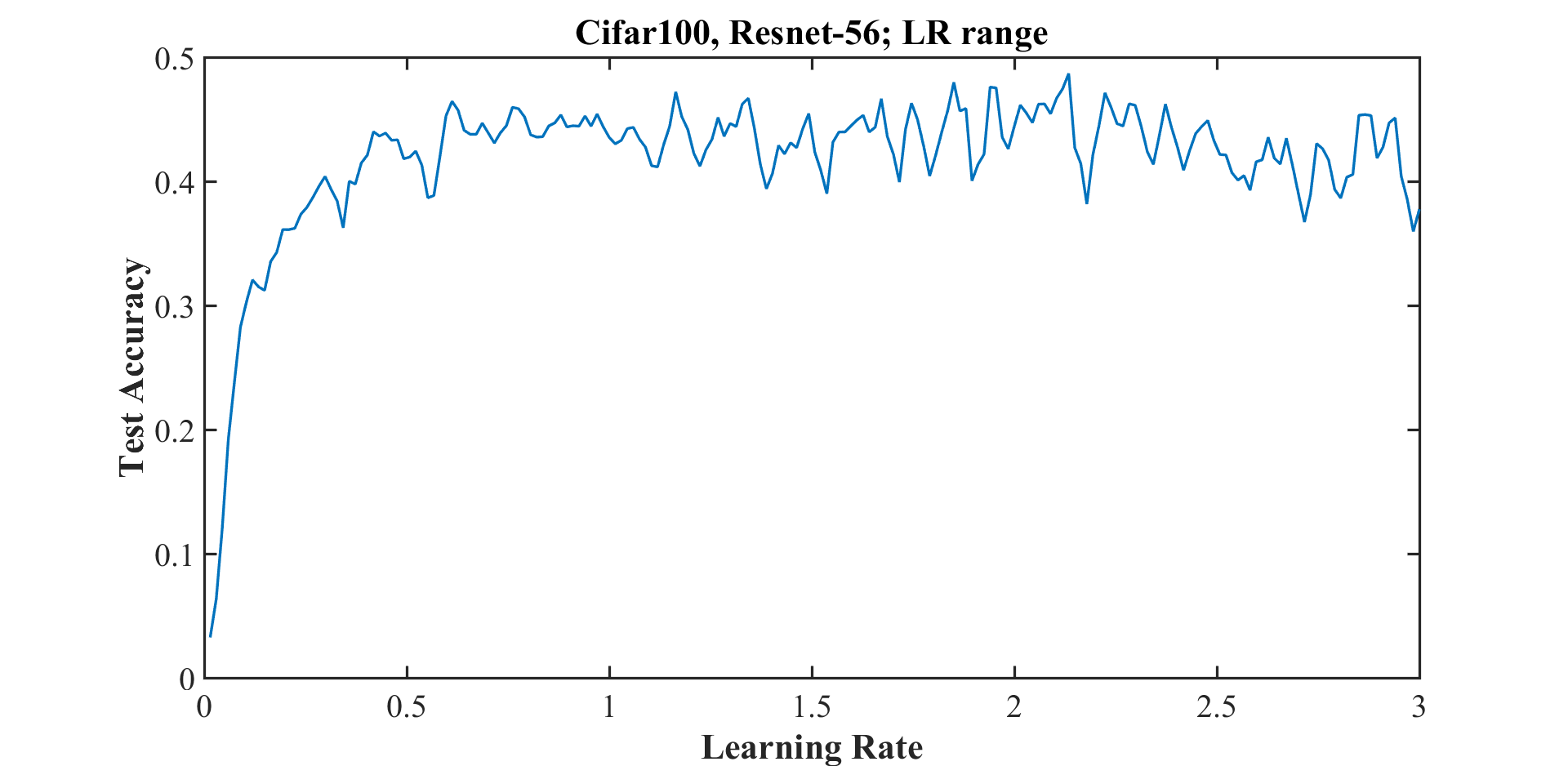}
			\caption{Learning rate range test.}
			\label{fig:C100range3Res56}
		\end{subfigure}
		\quad
		\hfill
		~ 
		\centering
		\centering
		\begin{subfigure}[b]{0.47\textwidth}
			\includegraphics[width=\textwidth]{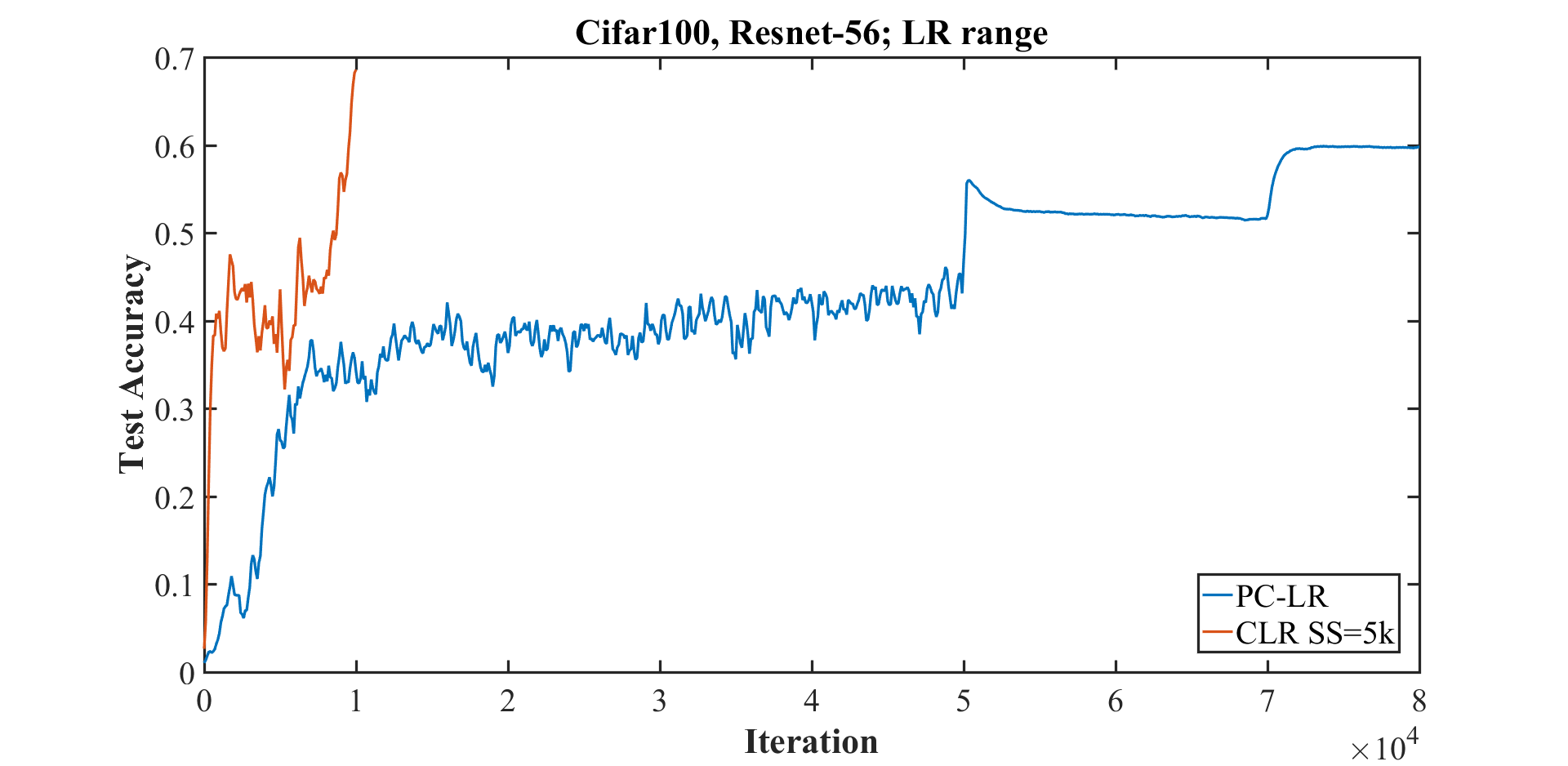}
			\caption{Comparison of test accuracies for super-convergence to a piecewise constant training regime.}
			\label{fig:c100res56CLRvsLR}
		\end{subfigure}
		\caption{Comparisons for Cifar-100, Resnet-56  of super-convergence to typical (piecewise constant) training regime. }
		\label{fig:Cifar100}
	\end{figure}

	\begin{figure} [tbh]
		\begin{subfigure}[b]{0.48\textwidth}
			\includegraphics[width=\textwidth]{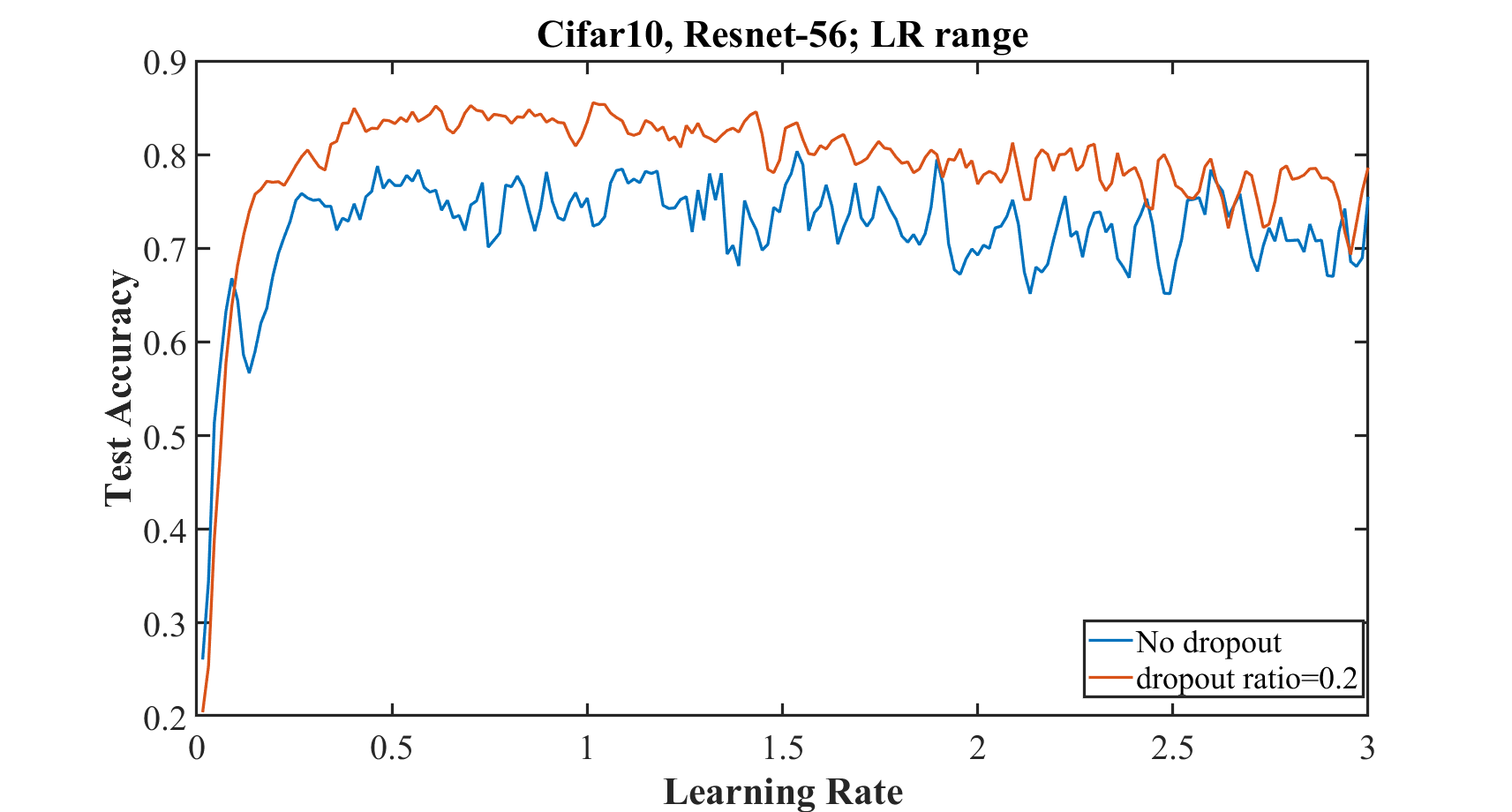}
			\caption{Comparison of LR range test for super-convergence with and without dropout.}
			\label{fig:lrRangeDropout}
		\end{subfigure}
		\quad
		\hfill
		~ 
		\centering
		\begin{subfigure}[b]{0.45\textwidth}
			\includegraphics[width=\textwidth]{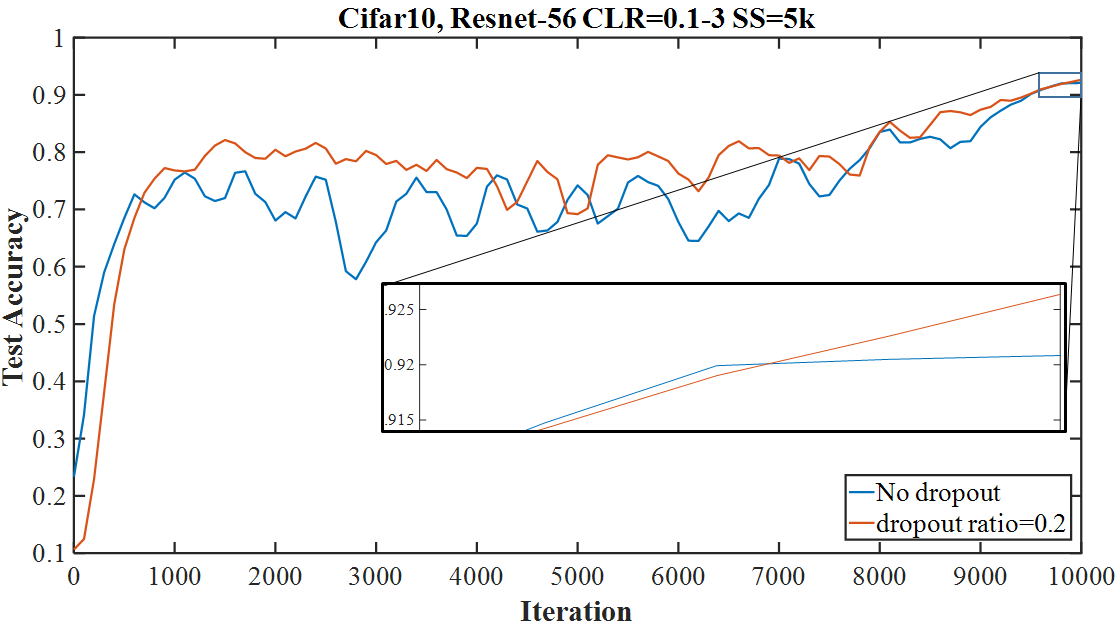}
			\caption{Comparison of test accuracies for super-convergence with and without dropout.}
			\label{fig:clr3SS5kRes56dropout123}
		\end{subfigure}
		\centering
		\caption{Comparisons of super-convergence with and without dropout (dropout ratio=0.2). }
		\label{fig:Dropout}
	\end{figure}
	
	\begin{figure} [tbh]
		\begin{subfigure}[b]{0.47\textwidth}
			\includegraphics[width=\textwidth]{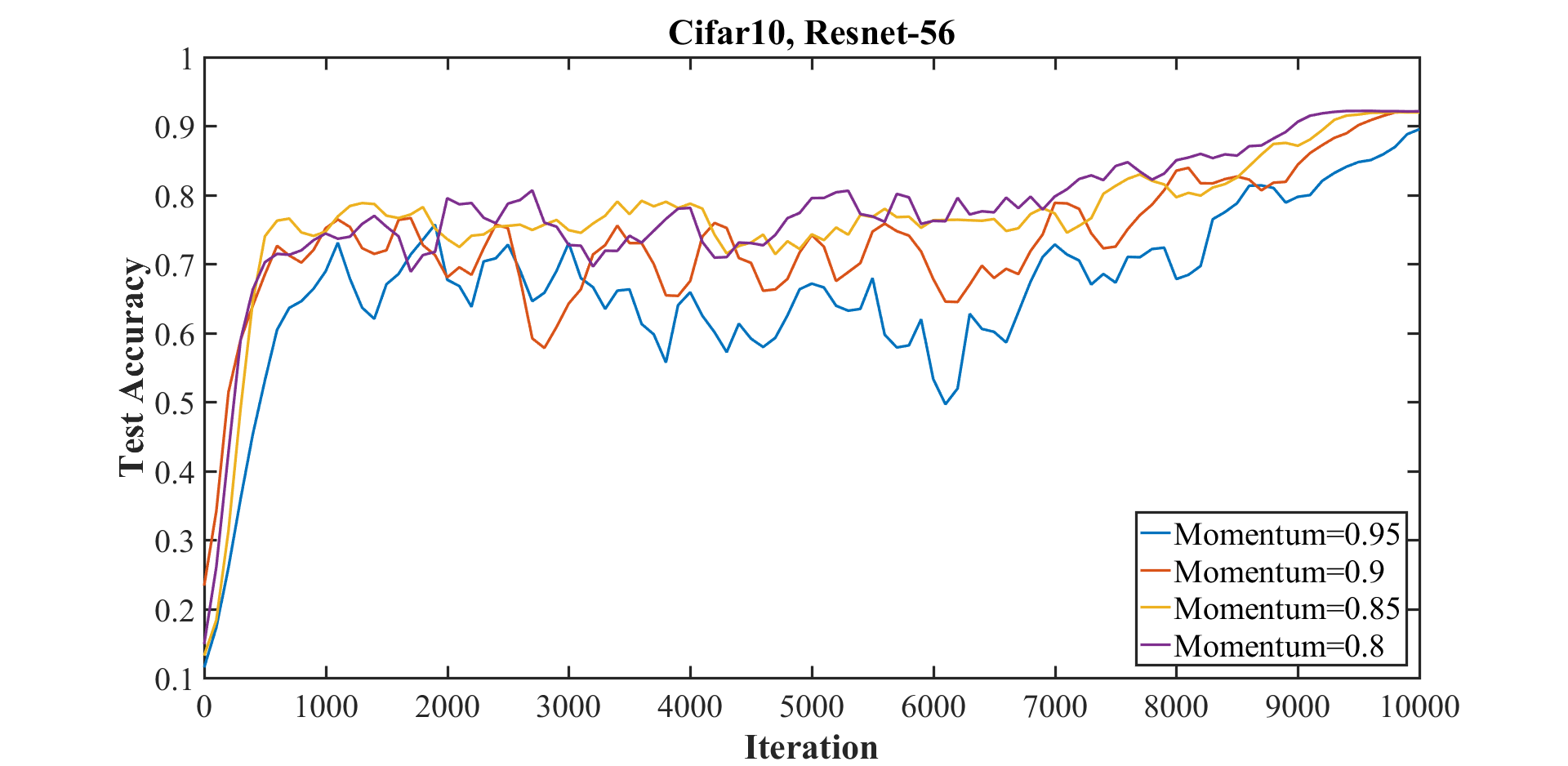}
			\caption{Comparison of test accuracies with various values for  momentum.}
			\label{fig:clr3SS10kRes56Mom}
		\end{subfigure}
		\quad
		\hfill
		~ 
		\centering
		\centering
		\begin{subfigure}[b]{0.47\textwidth}
			\includegraphics[width=\textwidth]{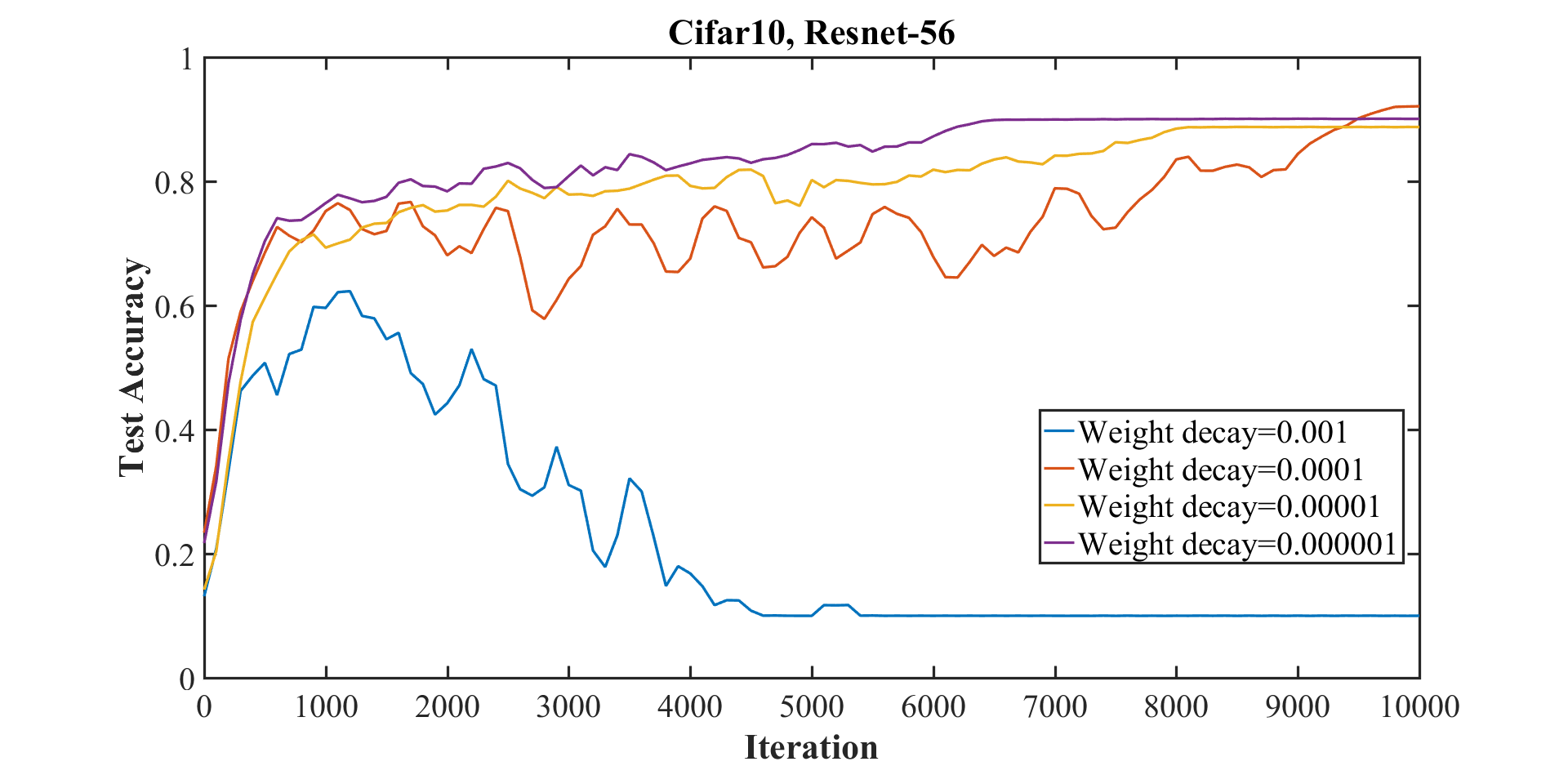}
			\caption{Comparison of test accuracies for super-convergence with various values of weight decay.}
			\label{fig:clr3SS5kResnet56WD}
		\end{subfigure}
		\caption{Comparisons for Cifar-10, Resnet-56  of super-convergence to typical training regime. }
		\label{fig:clr3SS5kother}
	\end{figure}
	
	
	\subsection{Experimental details and additional results}
	\label{sec:otherExp}
	
	We ran a wide variety of experiments and due to space limitations in the main article, we report results here that did not fit in the main article.
	
	The main paper shows final test accuracy results with Cifar-100, MNIST, and Imagenet, both with and without super-convergence.
	Figure \ref{fig:C100range3Res56} shows the results of the LR range test for Resnet-56 with the Cifar-100 training data.
	The curve is smooth and accuracy remains high over the entire range from 0 to 3 indicating a potential for super-convergence.
	An example of super-convergence with Cifar-100 with Resnet-56 is given in Figure \ref{fig:c100res56CLRvsLR}, where there is also a comparison to the results from a piecewise constant training regime.
	Furthermore, the final test accuracy for the super-convergence curve in this experiment is 68.6\%, while the accuracy for the piecewise constant method  is 59.8\%, which is an 8.8\% improvement.
	
	Figure \ref{fig:Dropout} shows a comparison of super-convergence, both with and without dropout.
	The LR range test with and without dropout is shown in Figure \ref{fig:lrRangeDropout}. 
	Figure \ref{fig:clr3SS5kRes56dropout123} shows the results of training for 10,000 iterations.
	In both cases, the dropout ratio was set to 0.2 and the Figure shows a small improvement with dropout.
	
	The effect of mini-batch size is discussed in the main paper but here we present a table containing the final accuracies of super-convergence training with various mini-batch sizes.
	One can see in Table \ref{tab1:results2} the final test accuracy results and this table shows that larger mini-batch sizes\footnote{GPU memory limitations prevented our testing a total batch size greater than 1,530} produced better final accuracies, which differs from the results shown in the literature.
	Most of results reported in the main paper are with a total mini-batch size of 1,000.


	\begin{figure} [tbh]
		\begin{subfigure}[b]{0.47\textwidth}
			\includegraphics[width=\textwidth]{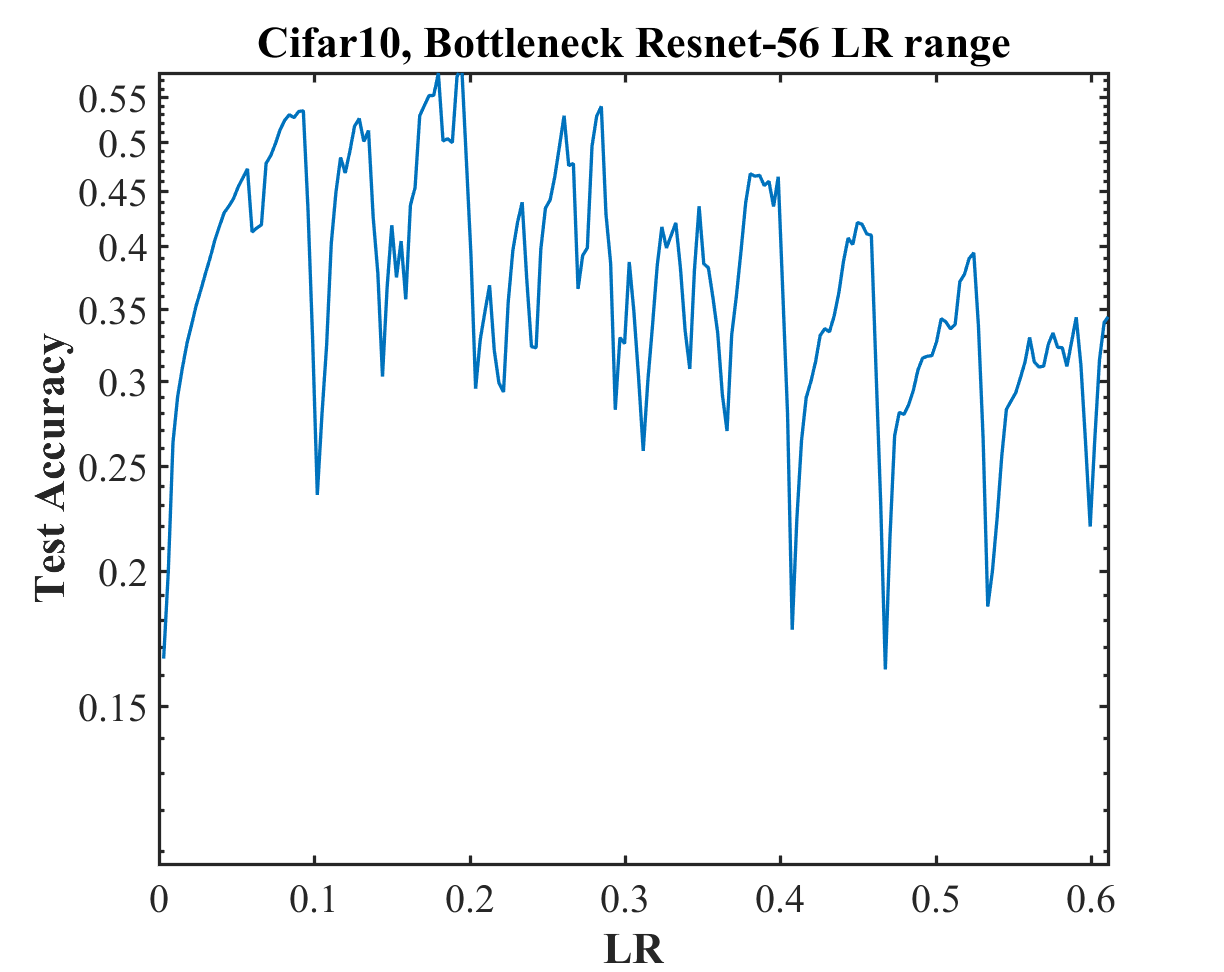}
			\caption{Learning rate range test result with the bottleneck Resnet-56 architecture.}
			\label{fig:ResNet56bnCifar10Range3}
		\end{subfigure}
		\quad
		\hfill
		~ 
		\centering
		\centering
		\begin{subfigure}[b]{0.47\textwidth}
			\includegraphics[width=\textwidth]{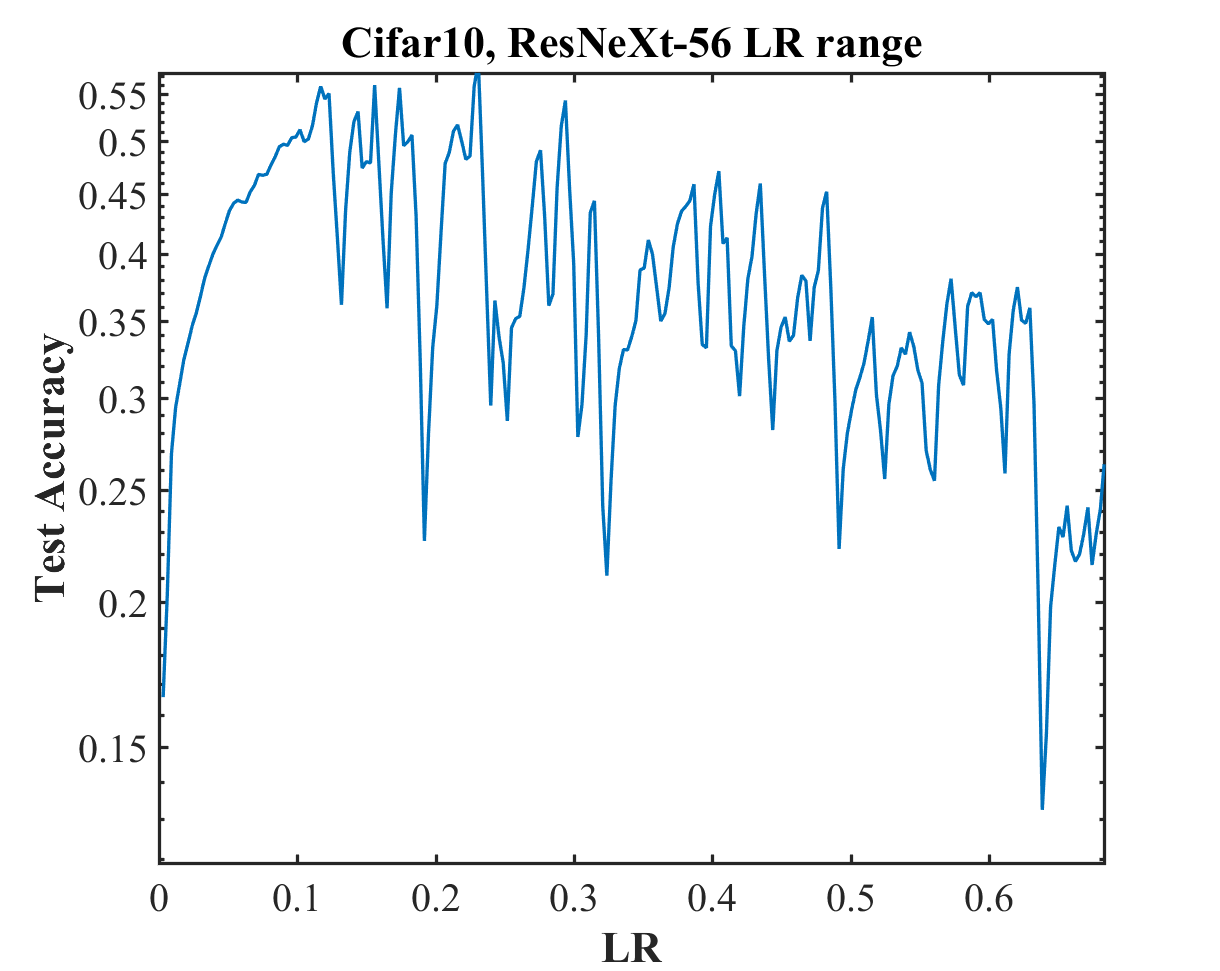}
			\caption{Learning rate range test result with the ResNeXt-56 architecture.}
			\label{fig:ResNeXt56Cifar10Range3}
		\end{subfigure}
		\caption{Comparison of learning rate range test results on Cifar-10 with alternate architectures.}
		\label{fig:LRrangeArch}
	\end{figure}

	\begin{table}[tb]
		\begin{center}
			\begin{tabular}{| c | c | c | c |}
				\hline
				Mini-batch size &  Accuracy (\%) \\ \hline \hline
				1536  & 92.1 \\ \hline
				1000  & 92.4 \\ \hline
				512  & 91.7 \\ \hline
				256  & 89.5 \\ \hline
				
			\end{tabular}
			\vspace{5pt}
			\caption{Comparison of accuracy results for various total training batch sizes with Resnet-56 on Cifar-10 using CLR=0.1-3 and stepsize=5,000.}
			\label{tab1:results2}
		\end{center}
		\vspace{-5pt}
	\end{table}
	
	\begin{table}[tb]
		\begin{center}
			\begin{tabular}{| c | c | c | c |}
				\hline
				Momentum &  Accuracy (\%) \\ \hline \hline
				0.80  & 92.1 \\ \hline
				0.85  & 91.9 \\ \hline
				0.90  & 92.4 \\ \hline
				0.95  & 90.7 \\ \hline
				
			\end{tabular}
			\vspace{5pt}
			\caption{Comparison of accuracy results for various momentum values with Resnet-56 on Cifar-10 using CLR=0.1-3 and stepsize=5,000.}
			\label{tab1:results3}
		\end{center}
		\vspace{-5pt}
	\end{table}
	
	


	In addition, we ran experiments on Resnet-56 on Cifar-10 with modified values for momentum and weight decay to determine if they might hinder the super-convergence phenomenon.
	Figure \ref{fig:clr3SS10kRes56Mom} shows the results for momentum set to 0.8, 0.85, 0.9, and 0.95 and the final test accuracies are listed in Table \ref{tab1:results3}.
	These results indicate only a small change in the results, with a setting of 0.9 being a bit better than the other values.
	In Figure \ref{fig:clr3SS5kResnet56WD} are the results for weight decay values of $ 10^{-3},  10^{-4},  10^{-5},$ and $10^{-6}$.
	In this case, a weight decay value of $ 10^{-3} $ prevents  super-convergence, while the smaller values do not.
	This demonstrates the principle in the main paper that regularization by weight decay must be reduced to compensate for the increased regularization by large learning rates.
	This Figure also shows that  a weight decay value of $ 10^{-4} $ performs well.

\end{document}